\documentclass[10pt,twocolumn,letterpaper]{article}

\usepackage{multirow}
\usepackage[pagenumbers]{cvpr} 
\usepackage{booktabs}
\usepackage{tcolorbox}

\definecolor{cvprblue}{rgb}{0.21,0.49,0.74}
\usepackage[pagebackref,breaklinks,colorlinks,allcolors=cvprblue]{hyperref}

\def\paperID{15255} 
\def\confName{CVPR}
\def\confYear{2025}

\title{Orient Anything: Learning Robust Object Orientation Estimation \\ from Rendering 3D Models}

\author{
   \vspace*{0.1cm}
   \!Zehan Wang$^{1}$\thanks{Equal Contribution.}\ , \  Ziang Zhang$^{1*}$,\  Tianyu Pang$^2$,\  Chao Du$^{2}$,\  Hengshuang Zhao$^3$,\  Zhou Zhao$^{1}$\\
   \vspace*{0.1cm}
   {$^1$}Zhejiang University; {$^2$}Sea AI Lab; {$^3$}The University of Hong Kong \\
   \url{https://orient-anything.github.io/}
}

\let\oldtwocolumn\twocolumn
\renewcommand\twocolumn[1][]{%
    \oldtwocolumn[{#1}{
    \begin{center}
    \vspace{-1\baselineskip}
    \includegraphics[width=1\textwidth]{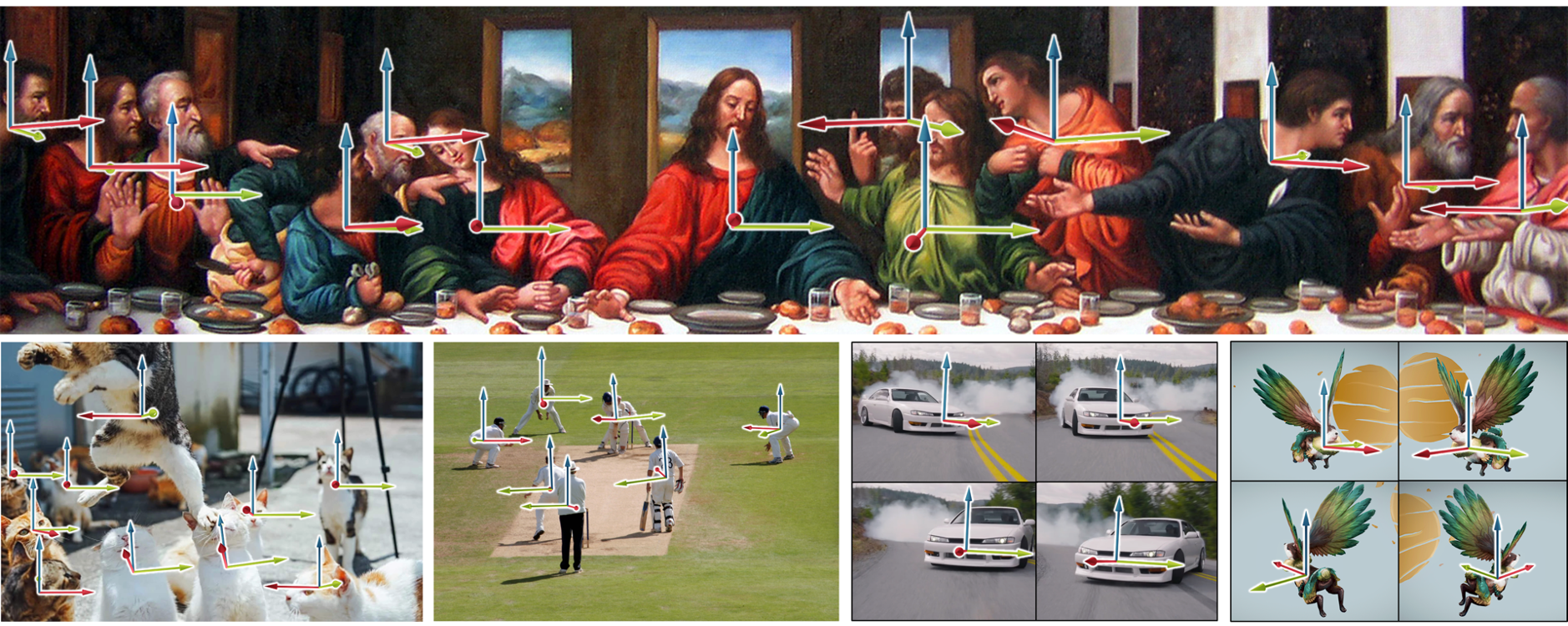}
    \vspace{-1\baselineskip}
    \captionof{figure}{We introduce a novel method for estimating the object orientation in images, represented by the red axis, while the blue and green axes indicate the upward and left sides of the object. More examples are provided in Appendix. \textit{Best viewed on screen with zoom.} 
    }
    \label{fig:case}
        \end{center}
    }]
}

\begin{document}
\maketitle

\begin{abstract}
Orientation is a key attribute of objects, crucial for understanding their spatial pose and arrangement in images. However, practical solutions for accurate orientation estimation from a single image remain underexplored. In this work, we introduce \textbf{Orient Anything}, the first expert and foundational model designed to estimate object orientation in a single- and free-view image. Due to the scarcity of labeled data, we propose extracting knowledge from the 3D world. By developing a pipeline to annotate the front face of 3D objects and render images from random views, we collect 2M images with precise orientation annotations. To fully leverage the dataset, we design a robust training objective that models the 3D orientation as probability distributions of three angles and predicts the object orientation by fitting these distributions. Besides, we employ several strategies to improve synthetic-to-real transfer. Our model achieves state-of-the-art orientation estimation accuracy in both rendered and real images and exhibits impressive zero-shot ability in various scenarios (Fig.~\ref{fig:case}). More importantly, our model enhances many applications, such as comprehension and generation of complex spatial concepts and 3D object pose adjustment.

\end{abstract}

\section{Introduction}
Perceiving object properties in a single image is the core problem in computer vision. Current visual foundation models and large vision-language models (VLMs) excel in tasks like object recognization~\cite{zhang2024recognize, liu2024visual}, localization~\cite{liu2023grounding, li2022grounded}, tracking~\cite{wang2023tracking, rajivc2023segment}, and segmentation~\cite{kirillov2023segment, ravi2024sam}. 

However, the object orientation, which is critical for understanding object pose and arrangement, has been underexplored due to the lack of annotated data. 
Omni3D~\cite{brazil2023omni3d} enables 3D orientation prediction by unifying 3D object detection data, but its scope is still restricted to specific domains, primarily room and street scenes, making it difficult to generalize to diverse real-world scenarios.

\begin{figure}[t]
    \centering
    \includegraphics[width=0.9\linewidth]{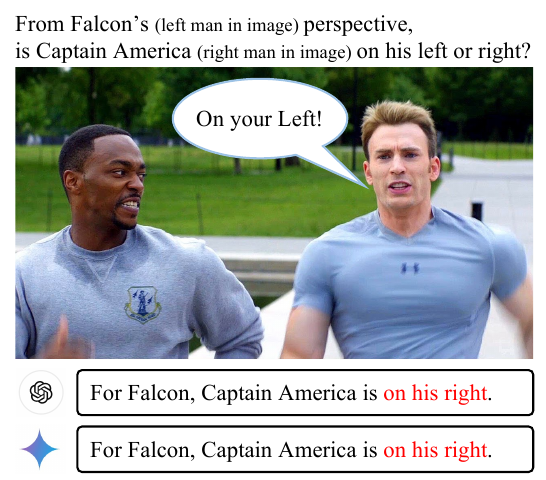}
    \vspace{-0.3\baselineskip}
    \caption{Understanding object orientation is essential for spatial reasoning. However, even advanced VLMs like GPT-4o and Gem- ini-1.5-pro are not yet able to resolve the basic orientation issue.}
    \vspace{-0.\baselineskip}
    \label{fig:gpt4}
\end{figure}

Furthermore, even the most advanced general visual understanding systems, like GPT-4o~\cite{hurst2024gpt} and Gemini~\cite{team2023gemini, team2024gemini}, struggle to comprehend basic object orientation. As a result, they perform poorly on questions derived from orientation, such as imagining object movement trends, or understanding object spatial relationships, as shown in Fig.~\ref{fig:gpt4}.

In this paper, we propose to learn how various objects look under different orientations by rendering 3D models. By annotating the front face of these 3D objects, we can easily and cheaply obtain precise orientation labels for each rendered view. This idea provides scalable, diverse, and easy-to-acquired data, enabling the development of accurate and generalizable orientation estimation models.

To this end, we develop a data collection pipeline to automatically filter, annotate, and render 3D assets~\cite{deitke2023objaverse}, enabling scalable data generation at any desired scale. In particular, we leverage advanced VLM~\cite{team2024gemini} to identify the front side of 3D objects from orthographic views, complemented by canonical pose detection and symmetry analysis to simplify the task and improve accuracy. Then, we render images from random perspectives, using azimuth and polar angles relative to the object orientation vector, combined with the camera rotation angle, to represent the 3D orientation.


Although scalable orientation data is available now, training a reliable orientation prediction model remains non-trivial. Direct regression of the three angles struggles to converge, resulting in poor performance. To overcome this challenge, we reformulate the single angle values as probability distributions to better capture the correlation between adjacent angles. By driving the model to fit these angle probability distributions, we simplify the learning process and significantly enhance model robustness.
Furthermore, considering the domain gap between the rendered and real images, we investigate various model initializations that incorporate real-world prior knowledge, alongside data augmentation strategies to improve synthetic-to-real transfer.

Our contribution can be summarized as:
\begin{itemize}
    \item We develop a reliable and automatic 3D object orientation annotation pipeline, and highlight the values of rendering 3D objects for generating cost-effective, diverse, and scalable image datasets with precise orientation labels.
    \item We introduce the orientation probability distribution fitting task as the learning objective to stabilize the training process and improve generalization.
    \item We investigate various model initialization and data augmentation strategies to improve synthetic-to-real transfer.
    \item Our model exhibits much stronger orientation estimation ability compared to both the expertise model (Cube RCNN) and leading VLMs (GPT-4o and Gemini).
\end{itemize}

\section{Related Work}

\subsection{Orientation-based Understanding} Object orientation provides context about how objects are positioned relative to one another and to the viewer (or the camera), which is fundamental for object pose and relationship understanding. Accurate orientation understanding plays a key role in many advanced applications.

In 3D scene understanding, many studies~\cite{chen2020scanrefer, achlioptas2020referit3d, azuma2022scanqa} have highlighted the importance of spatial relationships informed by object orientation. SQA3D~\cite{ma2022sqa3d} first describes the position and orientation of an agent in a 3D scene, then tasks the model with answering questions based on the given spatial context. EmboidedScan~\cite{wang2024embodiedscan} manually annotates orientations for 3D objects and utilizes the pose information to describe spatial relationships among objects in 3D space.

In the domain of 2D images, understanding object orientation is also fundamental for accurately interpreting~\cite{goral2024seeing, wu2024human} or generating~\cite{shi2023mvdream, huang2024orientdream, wei2023lego} spatial relationships and properties. \citet{goral2024seeing} propose the visual perspective-taking task to assess 2D VLM's ability to understand the orientation and viewpoint of a person in images and highlight various applications based on this ability. Furthermore, the object orientation relative to the camera determines its pose in the image, which is essential for distinguishing spatial properties such as the front wheels of cars and the left shoulder of a person, along with complex spatial relationships. Moreover, generating objects with given pose conditions is vital for controllable image generation~\cite{huang2024orientdream, wei2023lego}.

Although object orientation is closely linked to numerous questions and applications, practical solutions for estimating object orientation in images are still underexplored. Our work fills this gap by proposing the first foundation model for object orientation estimation, which exhibits strong zero-shot performance in real-world scenarios.

\subsection{Object Orientation Recognition in Images} Some tasks attempt to recognize object orientations in images under certain conditions or with extra information.

6DoF object pose estimation~\cite{guan2024survey} focuses on detecting the position and orientation of objects in images. However, existing methods require the CAD model of the target object~\cite{sundermeyer2018implicit, peng2019pvnet} or other reference view of this same object~\cite{goodwin2022zero, nguyen2024nope, fan2024pope}, which means that these methods cannot infer the orientation of the object from a single image. On the other hand, rotated object detection~\cite{xie2021oriented, han2021align, brazil2023omni3d} focuses on generating rotatable 2D or 3D bounding boxes for objects. Omni3D~\cite{brazil2023omni3d} unifies multiple 3D object detection datasets and trains Cube R-CNN to identify the 3D position and orientation of objects from a single image. While Cube R-CNN demonstrates a certain capability in detecting objects in 3D space, its performance is constrained by the data scope of Omni3D, which predominantly features indoor scenes and street environments. Furthermore, the orientation angles predicted by Cube R-CNN are primarily used to rotate the 3D bounding box while not always aligning with the front face of the objects. 


Unlike the aforementioned tasks, our work focuses on 3D orientation estimation of objects in single- and free-view images, and the orientation is strictly aligned with the meaningful front face of the objects.

\begin{figure*}[t]
    \centering
    \includegraphics[width=\linewidth]{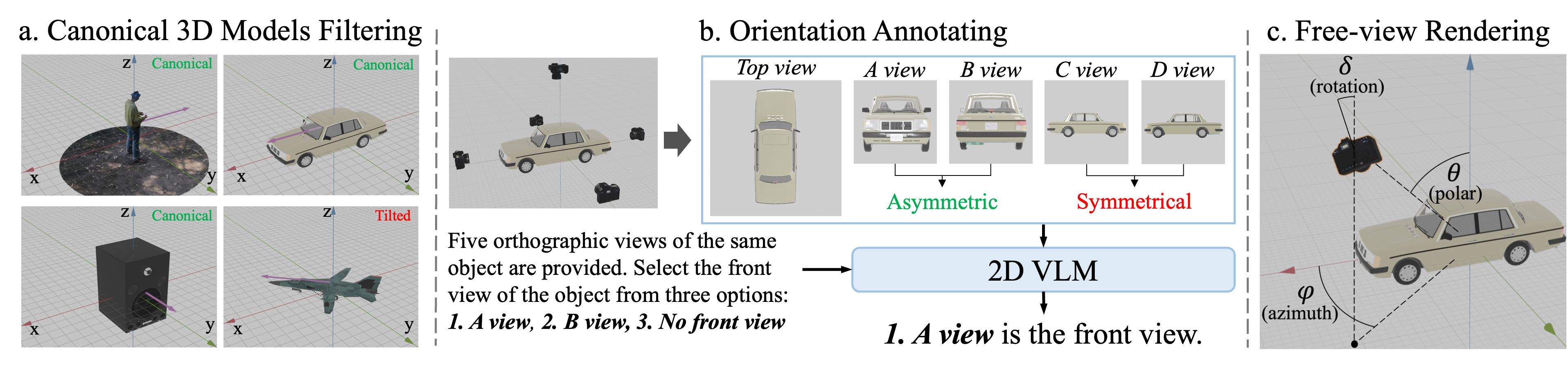}
    \vspace{-1.5\baselineskip}
    \caption{The orientation data collection pipeline is composed of three steps: \textbf{1) Canonical 3D Model Filtering}: This step removes any 3D objects in tilted poses. \textbf{2) Orientation Annotating}: An advanced 2D VLM is used to identify the front face from multiple orthogonal perspectives, with view symmetry employed to narrow the potential choices. \textbf{3) Free-view Rendering}: Rendering images from random and free viewpoints, and the object orientation is represented by the polar $\theta$, azimuthal $\varphi$ and rotation angle $\delta$ of the camera.}
    \label{fig:annotation}
\end{figure*}

\section{Orientation Understanding in 2D VLMs}
\label{sec:oribench}
Before proposing our method for object orientation estimation, we first investigate whether the ability to understand object orientation emerges in 2D VLMs trained on web-scale image datasets with billions of parameters.


To this end, we introduce Ori-Bench, the first VQA benchmark specifically designed to assess the capacity of 2D VLMs to understand object orientation and tackle related questions. We manually curate 200 images in total, with 100 from COCO~\cite{lin2014microsoft} and 100 generated by DALL-E 3~\cite{betker2023improving}. To substantively evaluate the understanding of object orientation, each image is horizontally flipped to produce a paired mirrored version, with answers adapted accordingly. A sample will be marked as solved only if the model correctly answers the question on both versions. There are three kinds of tasks: \textit{(1) Object Direction Recognition (73+73 samples): }identifying the orientation of an object within images; \textit{(2) Spatial Part Reasoning (39+39 samples): }distinguish parts of an object with specific spatial meanings, like left \textit{vs.} right hand of human; and \textit{(3) Spatial Relation Reasoning (88+88 samples): }imagining the relative position of one object from the perspective of another.

In Tab.~\ref{tab:oribench}, we show the accuracy of GPT-4o, Gemini-1.5-Pro, and our Orient Anything+LLM (Refer to Sec.~\ref{sec:understanding} for details). In the basic direction recognition task, the advanced VLMs can only correctly solve around 60\% samples. This limitation is especially evident in spatial reasoning and relation tasks, where the powerful GPT-4o and Gemini-1.5-Pro perform similarly to random guessing.
The pilot study highlights the need for fundamental tools to precisely estimate object orientation in images. All the examples are provided in Supplementary Materials.

\begin{table}[t]
\centering
\setlength\tabcolsep{8pt}
\resizebox{\linewidth}{!}{
\begin{tabular}{ccccc}
\toprule
 & \begin{tabular}[c]{@{}c@{}}Object\\ Direction\end{tabular} & \begin{tabular}[c]{@{}c@{}}Spatial\\ Part\end{tabular} &  \begin{tabular}[c]{@{}c@{}}Spatial\\ Relation\end{tabular} & Overall \\ \midrule
Random & 12.93 & 22.12 & 17.54 &16.75\\
GPT-4o & 49.32 & 15.38 & 27.27 & 32.50\\
Gemini-1.5-pro & 58.90 & 15.38& 18.18 & 33.00\\ \midrule
Orient Anything+LLM & \textbf{67.12} & \textbf{46.15} & \textbf{40.91} & \textbf{51.50}\\ \bottomrule
\end{tabular}}
\vspace{-0.6\baselineskip}
\caption{Quantitative results on the proposed Ori-Bench.}
\label{tab:oribench}
\vspace{-1\baselineskip}
\end{table}

\section{Orientation Data Collection}
The scarcity of orientation annotations is a major obstacle to learning general orientation estimation. Existing annotations for images, typically captions~\cite{schuhmann2021laion, schuhmann2022laion}, bounding boxes~\cite{lin2014microsoft}, or segmentation masks~\cite{kirillov2023segment, zhou2017scene}, seldom include object orientation information, and manually annotating object orientation in images is extremely time-consuming and costly. To overcome this limitation, we propose to utilize the 3D assets. Annotating the front face of 3D objects and then rendering images from random perspectives provides an efficient and effective way to generate large-scale image datasets with precise orientation annotations.

To this end, we first develop an automatic 3D object's orientation annotation and rendering pipeline, as shown in Fig.~\ref{fig:annotation}. Each step of the pipeline is detailed below.

\paragraph{Step1: Canonical 3D Models Filtering} We use Objaverse~\cite{deitke2023objaverse}, a large-scale dataset containing 800K object assets, as our database. Although most objects in this dataset are modeled in canonical poses (standing upright and facing one of four orthogonal directions along the $x$, $-x$, $y$, and $-y$ axes), some are tilted along orthogonal axes, as shown in Fig.~\ref{fig:annotation}.a.
To simplify orientation annotation and enhance reliability, we first exclude all the tilted assets, focusing solely on 3D objects in canonical poses. This idea reduces the 3D object orientation annotation problem to a multi-class classification task. Rather than identifying the specific orientation vector, we only need to determine the front face from images rendered alone $x$, $-x$, $y$, and $-y$ axes, or to conclude that the object has no front face.

To filter out tilted objects, we analyze the tilt in the three orthographic views of each object. Specifically, we extract the object edges for each view, and use Principal Component Analysis (PCA) to capture the principal directions of edges. If the principal edge direction is parallel with any coordinate axis (with a tolerance of two degrees for robustness) across all renderings, the object is considered to be in the canonical pose; otherwise, it will be deemed tilted.

Starting with the initial pool of 800K objects in the Objaverse dataset, we first curate 80K 3D models with high texture quality. Using our tilt-filtering criteria, we select 55K objects in canonical poses for subsequent processing.

\paragraph{Step2: Orientation Annotating} Using the selected 3D objects in canonical poses, we render four orthogonal views from the $x$, $-x$, $y$ and$-y$ axes, along with a top view for additional global reference. Although our pilot study in Sec.~\ref{sec:oribench} indicates that current 2D VLMs struggle to accurately predict orientation from a single view, we find that they perform well in identifying which view is facing the camera when multiple orthogonal views are presented for comparison and reference.

Additionally, to mitigate VLM hallucinations and improve annotation accuracy, we incorporate symmetry as auxiliary information. Since the front and back faces of objects are typically asymmetrical, we leverage this prior knowledge to further narrow down the possible choices. Specifically, we use a combination of SIFT~\cite{lowe2004distinctive}, structural similarity, and pixel color similarity to assess the similarity between opposing views. Two views are considered symmetrical if their similarity exceeds the threshold. Gemini-1.5-Pro is tasked with identifying the front face of objects from asymmetrical opposing views. If the object is symmetrical along both the $x$ \textit{vs.} $-x$ and $y$ \textit{vs.} $-y$, it is regarded as having no meaningful front face and orientation.

\paragraph{Step3: Free-View Rendering} Once the 3D object's orientation is annotated in 3D space, we can obtain its 3D orientation in images from any viewpoints. For simplicity and clarity, we use the spherical coordinate system to define object orientation.
As depicted in Fig.~\ref{fig:annotation}.c, we calculate the relative polar angle $\theta$ and azimuth angle $\varphi$ between the camera position and the object orientation axis, as well as the camera rotation angle $\delta$, to represent the object orientation from the specific viewpoint.

Before rendering, all 3D objects are scaled to a unit cube, with their centers aligned to the origin of the coordinate system. For each object, 40 images are rendered from random perspectives, with the camera aimed at the origin and each image rendered at 512$\times$512 resolution. In total, we collect 2M rendered images with precise orientation annotations.

\begin{figure*}[t]
    \centering
    \includegraphics[width=0.9\linewidth]{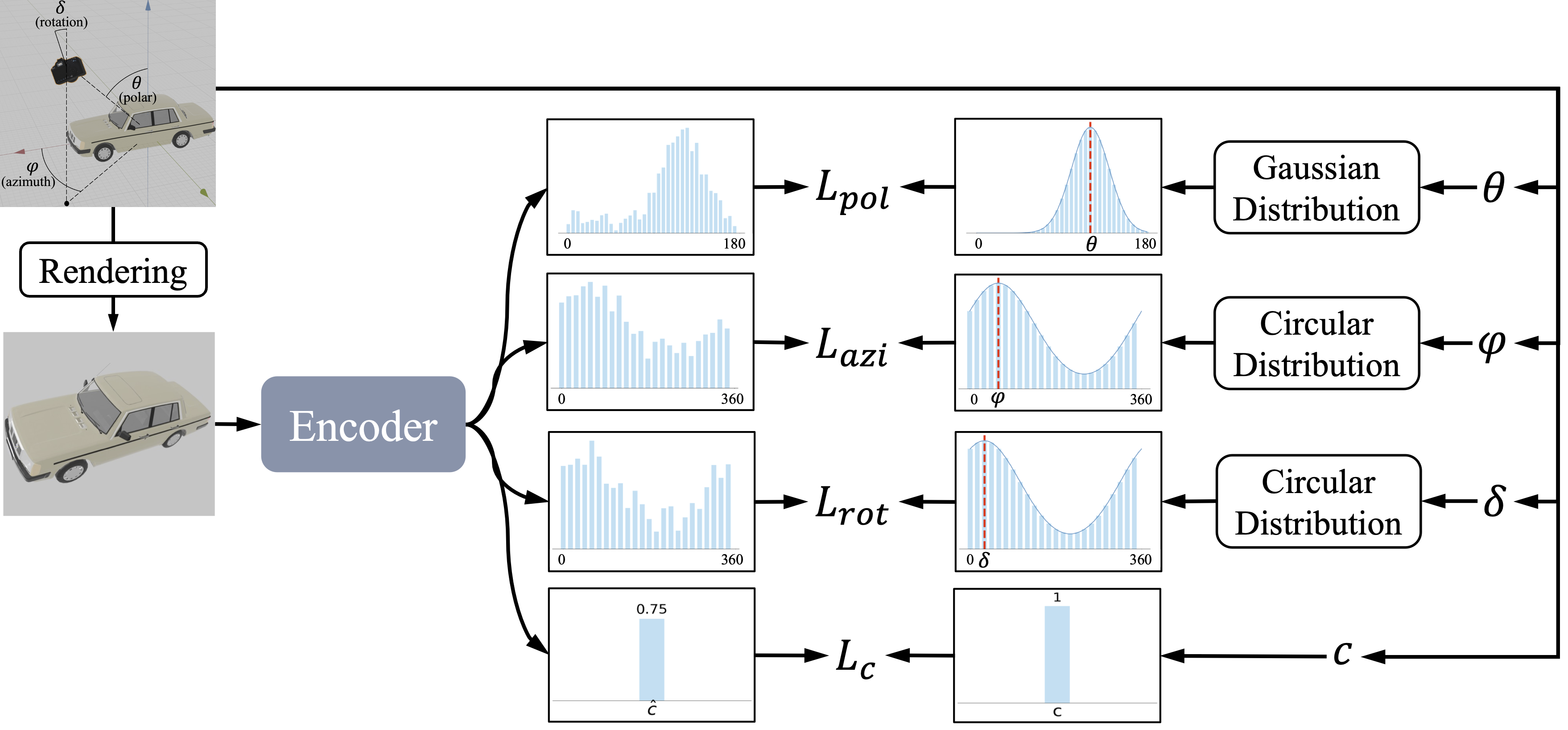}
    \caption{Orient Anything consists of a simple visual encoder and multiple prediction heads. It is trained to judge
    if the object in the input image has a meaningful front face and fits the probability distribution of 3D orientation.}
    \label{fig:Model}
\end{figure*}

\section{Orient Anything}
Based on the massive images of objects with annotated 3D orientation $\theta$, $\varphi$, and $\delta$, we train Orient Anything for general object orientation estimation in images.

\subsection{Orientation Probability Distribution Fitting}
\label{sec:fitting}
Despite having accurate 3D orientation annotations, developing an effective learning objective to guide accurate and robust orientation predictions is non-trivial. Our initial approach, which involved directly predicting continuous angle values with L2 loss as supervision, struggles to converge and performs poorly.

To address this, we first simplify the challenging continuous regression task into a discrete classification problem, which is easier to optimize. Specifically, we divide the 360° range into 360 individual classes, each representing a 1° interval. While lowering the task difficulty improves performance over continuous regression, it fails to capture correlations between adjacent angles that produce nearly identical outcomes in practice (e.g., rendering at polar 29°, 30°, and 31°). Treating these close angles as independent classes neglects their inherent relationships, which may confuse the model. Therefore, we further reformulate the classification task as a discrete probability distribution fitting problem, which is also easy to converge and can fully capture the potential relationship between different orientations.

\paragraph{Target Probability Distribution} We first transform the ground-truth angles into target probability distributions, represented as Gaussian distributions centered on the ground-truth angle, with manually set variances. These distributions are subsequently discretized into a grid-based format at 1° intervals. For a given ground-truth polar angle $\theta$ (in degrees), the probability distribution of polar angle $\mathbf{P}_{\textrm{pol}}(i|\theta, \sigma_\theta)$ can be formulated as follows:
\begin{equation}
\begin{aligned}
    \mathbf{P}_{\textrm{pol}}\left(i|\theta, \sigma_\theta\right) = \frac{\operatorname{exp}\left(-\frac{(i-\theta)^2}{2\sigma_\theta^2}\right)}{\sum_{n=1}^{180} \operatorname{exp}\left(-\frac{(n-\theta)^2}{2\sigma_\theta^2}\right)}\textrm{,}
\end{aligned}
\end{equation}
where $i=1^{\circ},\dots,180^{\circ}$ and $\sigma_\theta$ is the variance hyper-parameter for polar distribution. For the ground truth azimuth angle $\varphi$ and rotation angle $\gamma$, due to its periodicity (e.g., 359°, 360°, and 1° are adjacent), we employ the circular Gaussian distribution to from their target distribution $\mathbf{P}_{\textrm{azi}}(i | \varphi, \sigma_{\varphi})$ and $\mathbf{P}_{\textrm{rot}}(i | \delta, \sigma_{\delta})$. For brevity, we illustrate this process using azimuth as an example:
\begin{equation}
\mathbf{P}_{\textrm{azi}}\left(i | \varphi, \sigma_{\varphi}\right) = \frac{\operatorname{exp}\left({\frac{\cos(i - \varphi)}{\sigma_{\varphi}^2}}\right)}{2\pi I_0\left(\frac{1}{\sigma_{\varphi}^2}\right)}\textrm{,}
\end{equation}
where $i=1^{\circ},\dots,360^{\circ}$, $\sigma_\varphi$ is the variance for polar distribution, and $I_0(\frac{1}{\sigma_{\varphi}^2})$ is the zero-order modified Bessel function of the first kind, which can be represented as:
\begin{equation}
    I_0\left(\frac{1}{\sigma_{\varphi}^2}\right) = \sum_{n=0}^{\infty} \frac{1}{(n!)^2}\left(\frac{1}{2\sigma_{\varphi}^2}\right)^{2n}\textrm{.}
\end{equation}

As shown in Fig.~\ref{fig:Model}, the circular Gaussian distribution effectively models the periodicity of azimuth and rotation angles, ensuring the stability of the optimization process.

\paragraph{Training and Inference} Given the input image $I$, we use a visual encoder to extract its latent feature, followed by prediction heads (simple linear layers) to output the distributions of polar, azimuth and rotation angles: $\widehat{\mathbf{P}}_{\textrm{pol}} \in \mathbb{R}^{180}$, $\widehat{\mathbf{P}}_{\textrm{azi}} \in \mathbb{R}^{360}$, and $\widehat{\mathbf{P}}_{\textrm{rot}} \in \mathbb{R}^{360}$, respectively, representing the object orientation in 3D space. Additionally, the model predicts an orientation confidence $\hat{\mathbf{c}} \in \mathbb{R}^1$, to determine whether the object has a defined front face and orientation. This approach is used for handling centrally symmetric objects like balls and stools.
The target distributions $\mathbf{P}_{\textrm{pol}}(i | \theta, \sigma_{\theta})$, $\mathbf{P}_{\textrm{azi}}(i | \varphi, \sigma_{\varphi})$ and $\mathbf{P}_{\textrm{rot}}(i | \delta, \sigma_{\delta})$ are defined above, with the orientation label $\mathbf{c}$ being 1 if the object has a front face, and 0 otherwise. 
We use cross-entropy (CE) loss to supervise the predicted orientation distributions, and the corresponding loss terms are denoted as: $L_{\textrm{pol}}$, $L_{\textrm{azi}}$ and $L_{\textrm{rot}}$. For $\hat{\mathbf{c}}$, binary cross-entropy (BCE) loss is employed, yielding $L_{\textrm{c}}$.
The final training loss is a linear combination of the above four terms, and for objects without meaningful orientation, the $L_{\textrm{pol}}$, $L_{\textrm{azi}}$, $L_{\textrm{rot}}$ will be disabled:
\begin{equation}
    L=
    \begin{cases}
        \lambda L_{\textrm{c}}, &\mathbf{c}=0 \\
        L_{\textrm{pol}}+ L_{\textrm{azi}} + L_{\textrm{rot}} + \lambda L_{\textrm{c}}, &\mathbf{c}=1
    \end{cases}
    \label{eq:loss}
\end{equation}
where $\lambda$ is the loss coefficient for orientation judgment. 

During the inference process, objects whose orientation confidence is lower than 0.5 would be thought to have no meaningful front face and orientation. Otherwise, the angles with the highest probability in each distribution: $\widehat{\mathbf{P}}_{\textrm{pol}}$, $\widehat{\mathbf{P}}_{\textrm{azi}}$, $\widehat{\mathbf{P}}_{\textrm{rot}}$ are taken as the predicted polar, azimuth, and rotation angle: $\hat{\theta}$, $\hat{\varphi}$, $\hat{\delta}$.

\begin{table*}[t]
\centering
\setlength\tabcolsep{4pt}
\begin{tabular}{cccccccccc}
\toprule
\multirow{4}{*}{Models} & \multicolumn{7}{c}{\textbf{Rendered Image}} & \multicolumn{2}{c}{\textbf{Real Image}} \\
\cmidrule(lr){2-8}\cmidrule(lr){9-10}
 & \textit{Judgment} & \multicolumn{2}{c}{\textit{Azimuth Estimation}} & \multicolumn{2}{c}{\textit{Polar Estimation}} & \multicolumn{2}{c}{\textit{Rotation Estimation}} & \textit{Judgment} & \textit{Recognition} \\
 \cmidrule(lr){2-2}\cmidrule(lr){3-4}\cmidrule(lr){5-6}\cmidrule(lr){7-8}\cmidrule(lr){9-9}\cmidrule(lr){10-10}
 & Acc$\uparrow$ & Abs$\downarrow$ & Acc@22.5°$\uparrow$ & Abs$\downarrow$ & Acc@5°$\uparrow$ & Abs$\downarrow$ & Acc@5°$\uparrow$ & Acc$\uparrow$ & Acc$\uparrow$ \\
 \midrule
Random          & 50.00 & - & 12.50 & - & 5.55  & - & 16.67 & 50.00 & 12.50 \\
Cube RCNN       & -     & 89.00 & 12.44 & 27.99 & 10.37 & 132.74 & 2.50  & -     & 20.25\\ 
Gemini-1.5-pro  & 57.29 &79.51  & 19.06 &20.10  & 16.31 &2.61  & 85.12 & 66.96 & 31.95 \\ 
GPT-4o          & 61.85 & 81.07 & 19.94 & 16.02 & 17.56 & 4.65 & 81.00 & 69.29 & 45.78 \\ \midrule
Ours (ViT-S) & 73.88 & 45.27 & 63.18 & 5.12 & 71.62 & 0.82 & 97.06 & 78.54 & 63.44\\ 
Ours (ViT-B) &74.88  &39.03  &71.94  & 3.81 &81.37  & \textbf{0.26} &\textbf{99.56}  &\textbf{81.25}  &70.19 \\ 
Ours (ViT-L) & \textbf{76.00} & \textbf{38.60} & \textbf{73.94} & \textbf{2.94} & \textbf{86.75} & 0.70 & 98.31 & 80.30 & \textbf{72.44}\\ \bottomrule
\end{tabular}
\vspace{-0.6\baselineskip}
\caption{Orientation estimation on both in-domain rendered images and out-of-domain real images. The best results are \textbf{bold}.}
\label{tab:main}
\end{table*}

\begin{table*}[t]
\setlength\tabcolsep{2pt}
\centering
\resizebox{1\linewidth}{!}{
\begin{tabular}{cccccccccccccccc}
\toprule
 & \multicolumn{3}{c}{\textbf{SUN RGB-D}}& \multicolumn{3}{c}{\textbf{KITTI}}& \multicolumn{3}{c}{\textbf{nuScenes}}& \multicolumn{3}{c}{\textbf{Objectron}} & \multicolumn{3}{c}{\textbf{ARKitScenes}} \\
 \cmidrule(lr){2-4} \cmidrule(lr){5-7} \cmidrule(lr){8-10} \cmidrule(lr){11-13} \cmidrule(lr){14-16}
 &\textit{Azimuth} & \textit{Polar} & \textit{Rotation} & \textit{Azimuth} & \textit{Polar} & \textit{Rotation} & \textit{Azimuth} & \textit{Polar} & \textit{Rotation}& \textit{Azimuth} & \textit{Polar} & \textit{Rotation}& \textit{Azimuth} & \textit{Polar} & \textit{Rotation}  \\ \midrule

Cube RCNN & 93.58 &	39.73 &	140.10  &98.61 &	39.73 &	121.21& 89.63 &	15.64 	&132.57& 122.99 &	60.01 &	113.31 &91.16 &	37.39 &	132.86   \\ \midrule
Ours (ViT-S) & 58.20 &	11.63 &	\textbf{3.59}& 65.85 &	5.00 	&1.08 &72.68 &	5.58 	&2.16 &39.45 	&23.47& 	18.26 &69.37& 	14.25 &	2.63   \\
Ours (ViT-B) & 56.34 &	9.15 &	3.75 &54.02 &	5.86 &	\textbf{0.21} &66.56 &	5.72 &	\textbf{1.28} &36.49 &	\textbf{22.13} &	\textbf{18.34}& 75.45 &	12.48 &	\textbf{2.60}  \\
Ours (ViT-L)& \textbf{42.98} &	\textbf{8.38} 	&3.66 &\textbf{44.22} &	\textbf{3.57} &	0.89 &\textbf{55.17} &	\textbf{4.08} &	1.78& \textbf{30.09} &	22.19& 	18.54& \textbf{67.56}& 	\textbf{11.47}& 	2.82  \\ \bottomrule
\end{tabular}}
\vspace{-0.6\baselineskip}
\caption{Zero-shot orientation estimation on five unseen real image benchmarks. Reported in absolute error.}
\label{tab:multiple}
\end{table*}

\subsection{Sythetic-to-Real Transferring}
Although the rendered images of 3D objects provide extensive data with orientation annotations, there is a distribution shift between synthetic rendered images and real images. We try to prompt effective synthetic-to-real transfer from two aspects: integrating real-world pre-training knowledge and narrowing the training-inference domain gap.

\paragraph{Inheriting Real-world Knowledge by Initialization} As demonstrated in~\cite{yang2024depth, ke2024repurposing}, initializing the model with strong visual encoders pre-trained on real images can significantly improve its synthetic-to-real transfer ability. To evaluate this in our orientation estimation task, we train models initialized from 3 widely-used image pre-trained encoders: MAE~\cite{he2022masked}, CLIP~\cite{radford2021learning}, and DINOv2~\cite{oquab2023dinov2}. After trials and failures, DINOv2 yields satisfactory results, attributed to its task-agnostic pre-training, fine-grained perception, and strong generalization capabilities. Consequently, we develop our model using DINOv2 initialization.

\paragraph{Narrowing Domain Gap by Data Augmentation}
There are two main differences between rendered and real images. We employ corresponding data augmentation strategies to reduce the domain gap and enhance transfer performance.

First, objects in rendered images are typically fully visible, whereas real-world images often contain partially visible or occluded objects. To bridge this gap, we incorporate random cropping as a training data augmentation strategy. This technique simulates the occlusion situation in real-world images, thereby improving the model’s ability to generalize to real-world scenarios.

Second, to avoid ambiguity, the rendered image contains only one object. In contrast, real-world images often feature multiple objects. To adapt our model for such cases, we isolate each object using segmentation masks and estimate their orientations individually. This approach replicates the style of rendered images, broadens the applicability of our model, and enhances its performance on real-world images.

\section{Experiments}

\subsection{Implementation Details} We train models at three scales for different purposes: ViT-S, ViT-B, and ViT-L, all initialized with DINOv2. The loss coefficient $\lambda$ in Eq.~\ref{eq:loss} is set to 1. The variance hyperparameters $\sigma_\theta$, $\sigma_\varphi$, and $\sigma_\delta$ are configured as 2.0°, 20.0°, and 1.0°. For optimization, we use the AdamW~\cite{loshchilov2017decoupled} optimizer, with a learning rate of 1e-5 for the pre-trained visual encoder and 1e-3 for the newly introduced prediction heads. The models are trained for 50,000 steps with a batch size of 64 on the curated 2M object orientation dataset. All trainings are conducted on 4 A100 (40GB) GPUs.

\subsection{Rendered-Images Orientation Estimation}
\label{sec:render_image}
We first quantitatively validate the model by accurately estimating the numerical 3D orientation of the in-domain rendered images. We manually select and annotate 300 objects from Objaverse, of which 150 have orientation annotations and 150 have no meaningful front face and orientation. For each object, we render 16 images of random views, and there are 4,800 images for testing in total.

We evaluate methods from two aspects: \textit{1) Orientation Judgment}: Determine if the object has a meaningful front face. \textit{2) Orientation (azimuth, polar, rotation) Estimation}: Predict the accurate azimuth, polar, and camera rotation angles for objects, using Absolute Error (in degrees) and Acc@X° (accuracy within tolerances of $\pm$X°) as metrics. 
Expert image-based 3D object detection model, Cube RCNN~\cite{brazil2023omni3d} and advanced VLMs, GPT-4o~\cite{hurst2024gpt} and Gemini-1.5-pro~\cite{team2024gemini}, as used as baselines.

The results presented in Tab.~\ref{tab:main} showcase the superior performance of our model in accurately predicting 3D orientation for objects. In practical azimuth estimation, our method achieves more than triple the accuracy of previous approaches. Notably, the performance of Cube RCNN and advanced VLMs is only slightly better than random guessing, with a success rate of 19.94\% compared to 12.50\%. In contrast, the Orient Anything ViT-L yields 73.94\% accuracy and much lower absolute error, highlighting its practical value in reliably distinguishing object direction.

\subsection{Zero-shot Real-Image Orientation Recognition}
The primary goal of this work is to estimate object orientations in real images. To assess the model performance in real-world scenarios, we construct two kinds of evaluation benchmarks.

1) For objects in the wild, we collect objects from the COCO dataset and manually annotate their orientations. Given the difficulty of annotating accurate 3D orientation, we narrow our focus to more feasible scenarios by labeling object orientations on the horizontal plane in eight directions: front, back, left, right, front-left, front-right, back-left, and back-right. From the 80 categories available in the COCO validation set, we chose 20 images per category, creating a comprehensive benchmark of 1,600 samples in total. Two tasks are used for evaluation: \textit{(1) Orientation Judgment:} Determine whether the object has a front face and orientation. \textit{(2) Horizontal Direction Recognition:} Identify which of the eight directions on the horizontal plane the object is orienting and the recognition accuracy is reported.

2) For objects in room and street scenes, we conducted quantitative experiments on five real-world datasets: SUN RGB-D~\cite{song2015sun}, KITTI~\cite{geiger2012kitti}, nuScenes~\cite{caesar2020nuscenes}, Objectron~\cite{ahmadyan2021objectron} and ARKitScenes~\cite{baruch2021arkitscenes}. For each benchmark, 1,000 objects with 3D orientation annotations are randomly selected and cropped from the real images to form an orientation estimation benchmark. We assess the Cube RCNN (trained on in-domain data) and Orient Anything (trained on out-of-domain rendering data) by calculating the absolute error on azimuth, polar, rotation angles between predicted 3D orientation and ground truth.

As shown in Tab.~\ref{tab:main} and \ref{tab:multiple}, despite never being exposed to real-world images during training, each version of Orient Anything demonstrates clear superiority over existing alternative methods in recognizing object orientations in real images. We provide detailed results for each of the 80 object categories in the Appendix. Orient Anything consistently outperforms previous approaches by a significant margin across most categories, achieving over 90\% accuracy in major categories such as humans, animals, vehicles, and furniture.

\begin{table}[]
\resizebox{\linewidth}{!}{
\begin{tabular}{cccc}
\toprule
 & \begin{tabular}[c]{@{}c@{}}Single\\ View\end{tabular} & \begin{tabular}[c]{@{}c@{}}Canonical\\ Views\end{tabular} & \begin{tabular}[c]{@{}c@{}}Canonical\&\\ Symmetrical\end{tabular} \\ \midrule
 Gemini-1.5-pro & 44.00 & 74.00 & \textbf{86.00} \\
GPT-4o & 31.00 & 87.00 & \textbf{92.00} \\ \bottomrule
\end{tabular}}
\vspace{-0.6\baselineskip}
\caption{Ablation study for Orientation Annotation.}
\label{tab:ori}
\end{table}

\begin{figure}[t]
    \centering
    \includegraphics[width=\linewidth]{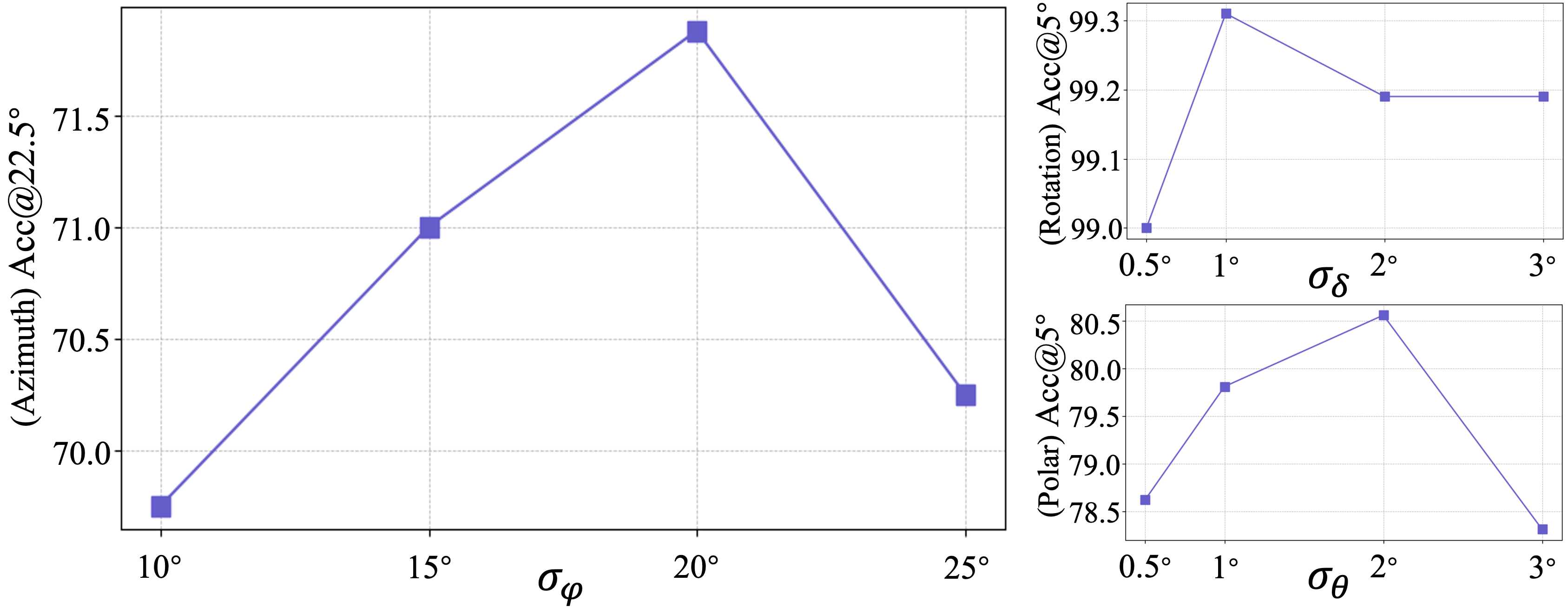}
    \vspace{-1.5\baselineskip}
    \caption{Ablation study for hyper-parameter $\sigma_\theta$, $\sigma_\varphi$ and $\sigma_\delta$.}
    \vspace{-1\baselineskip}
    \label{fig:sigma}
\end{figure}

Due to similar definitions and the same random guess results, ``Acc@22.5°" for rendered image azimuth estimation and ``Acc" for real image horizontal direction recognition are comparable. Our models achieve similar results on both metrics, highlighting the excellent synthetic-to-real transfer performance. For VLMs, recognizing horizontal directions in words is more accurate than predicting precise azimuth values in numbers, which reveals the shortcomings of VLMs in predicting precise values for 3D orientation. Cube RCNN, which predicts orientation values, performs significantly worse on rendered images due to its limited generalization capability.

We visualize our model predictions in Fig.~\ref{fig:case}, \ref{fig:generation}, and more in Appendix. These qualitative results highlight Orient Anything's remarkable zero-shot capability across images captured or created by real cameras, human artists, or generative models, as well as a variety of scenarios, including continuous video frames, multi-view images, and complex scenes containing multiple objects.


\begin{figure*}[t]
    \centering
    \includegraphics[width=0.95\linewidth]{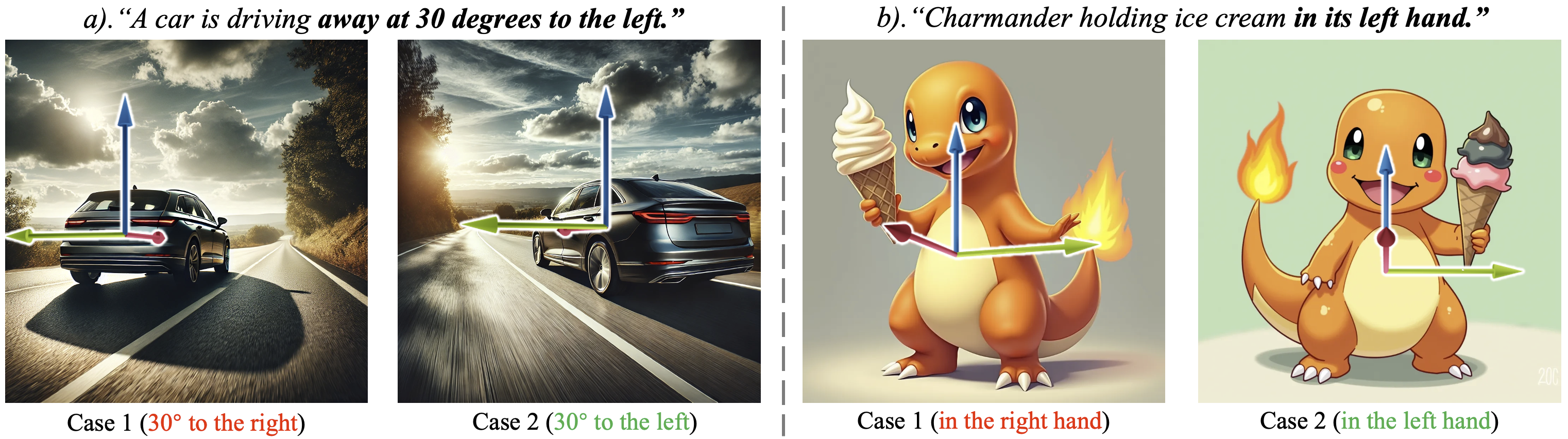}
    \vspace{-0.8\baselineskip}
\caption{Generated images with given textual prompt (left two from DALL-E 3~\cite{betker2023improving}, right two from FLUX~\cite{flux}). Accurate orientation estimation is helpful to confirm whether generated contents follow the given orientation or perspective condition.}
    \label{fig:generation}
\end{figure*}

\subsection{Ablation Study}

To verify the effectiveness of our key designs, we conduct ablation experiments using the ViT-B encoder.

\paragraph{Designs in Orientation Annotating} We evaluate our orientation annotating methods using the 300 manually annotated 3D objects introduced in Sec.~\ref{sec:render_image}. The results in Tab.~\ref{tab:ori} indicate that while VLMs achieve only 44\% and 31\% accuracy when identifying object orientation from the top view alone, providing orthogonal perspectives substantially enhances performance. Furthermore, incorporating symmetry as an extra condition further raises accuracy to nearly 90\%, underscoring the effectiveness of our orientation annotating strategy and proving the reliability of the rendering data.

\paragraph{Effect of $\sigma_{\theta}$, $\sigma_{\varphi}$ and $\sigma_{\delta}$} Fig.~\ref{fig:sigma} shows the effect of variance hyper-parameter for three kinds of angle probability distribution. In general, our method is insensitive to the variance selection, while most configurations yield superior results compared to the one-shot label.

\begin{table}[]
\setlength\tabcolsep{2pt}
\renewcommand{\arraystretch}{1.05}
\resizebox{\linewidth}{!}{
\begin{tabular}{ccccc}
\toprule
\multirow{3}{*}{Design} & \multirow{3}{*}{Variants} & \multicolumn{2}{c}{\textbf{Rendering Image}} & \textbf{Real Image} \\
\cmidrule(lr){3-4}\cmidrule(lr){5-5}
& & \textit{Azimuth} & \textit{Polar} & \textit{Recognition} \\
\cmidrule(lr){3-3}\cmidrule(lr){4-4}\cmidrule(lr){5-5}
& & Acc@22.5° &  Acc@5° & Acc \\ \midrule
\multirow{3}{*}{\begin{tabular}[c]{@{}c@{}}Learning\\ Objective\end{tabular}} 
&Regression& 12.00          & 20.50          & 21.48          \\
&Classification& 68.75          & 79.00          & 66.93          \\
& Fitting & \textbf{71.88} & \textbf{80.56} & \textbf{69.85} \\
\midrule
\multirow{5}{*}{\begin{tabular}[c]{@{}c@{}}Number\\ of Views\end{tabular}} 
&10 & 67.19 & 78.19 & 63.67         \\
&20 & 67.94 & 78.88 & 65.47         \\
&30 & 70.06 & 78.13 & 68.62         \\
&40 & \textbf{71.88} & 80.56 & \textbf{69.85}\\ 
&80 & 69.12& \textbf{80.69} &66.48\\ \midrule
\multirow{3}{*}{\begin{tabular}[c]{@{}c@{}}Training\\ Initialization\end{tabular}} 
&CLIP   & 58.44 & 71.88 & 49.27  \\
&MAE    & 58.44 & 64.63 & 57.26  \\
&DINOv2 & \textbf{71.88} & \textbf{80.56} & \textbf{69.85}\\ 
\midrule
\multirow{2}{*}{\begin{tabular}[c]{@{}c@{}}Training \\ Augmentation\end{tabular}} 
& None &71.88  &80.56  &69.85   \\
& Cropping & \textbf{71.94} & \textbf{81.37} & \textbf{70.19} \\ \midrule
\multirow{2}{*}{\begin{tabular}[c]{@{}c@{}}Inference \\ Augmentation\end{tabular}} 
& Box  & 71.88 & 80.56 & 67.49    \\
& Mask  & \textbf{71.88} & \textbf{80.56} & \textbf{69.85}   \\
 \bottomrule
\end{tabular}}
\vspace{-0.6\baselineskip}
\caption{Ablation study for Learning Objective, Number of Views, Training Initialization and Data Augmentation.}
\label{tab:ablation}
\vspace{-0.6\baselineskip}
\end{table}

\paragraph{Effect of Probability Prediction} In Tab.~\ref{tab:ablation}, we ablate the three learning objectives discussed in Sec.~\ref{sec:fitting}: continuous value regression, discrete angle classification, and probability distribution fitting. Direct regression yields poor performance, while angle classification performs significantly better but remains suboptimal. The final proposed probability distribution fitting method surpasses the alternatives, achieving markedly superior performance.


\paragraph{Number of Rendering Views} We explore the effect of the number of images rendered pre-3D object in Tab.~\ref{tab:ablation}. For a fair comparison, we train models to converge for each setting. The results indicate that too few views fail to provide sufficient information about objects from different perspectives, while overly dense sampling results in redundant images within the dataset, potentially hindering convergence.
Empirically, rendering 40 views for each object achieves the best balance and yields the optimal results.


\paragraph{Effect of Model Initialization} We compare several powerful pre-trained visual encoders as initialization for our orientation estimation task in Tab.~\ref{tab:ablation}. We empirically find that DIONv2 exhibits much better performance in both in-domain convergence and out-of-domain transfer compared to others, which may be attributed to its large-scale task-agnostic pre-training and superior fine-grained perception.

\paragraph{Effect of Data Augmentation} Tab.~\ref{tab:ablation} present the effect of data augmentation for improving sythetic-to-real transfer. During training, random cropping enables rendered images to mimic the objects occlusions, which significantly enhances the performance in real-world scenarios. For inference, using segmentation masks to isolate objects aligns more closely with the style of rendered images compared to bounding boxes, thereby narrowing the domain gap and improving overall performance.

\section{Applications}

\subsection{Spatial Understanding}
\label{sec:understanding}
Orientation is a key attribution for accurately understanding the spatial relations, as we highlighted in Sec.~\ref{sec:oribench} and Fig.~\ref{fig:gpt4}. We find that using Grounded-SAM~\cite{ren2024grounded} and our Orient Anything to identify object position and orientation in images and conveying these spatial details in pure text to an LLM~\cite{hurst2024gpt}, effectively addresses more orientation-based questions that confuse GPT-4o and Gemini-1.5-pro, as shown in Tab.~\ref{tab:oribench} and examples in Appendix. These results underscore the value of our model in spatial understanding.

\subsection{Spatial Generation Scoring} As shown in Fig.~\ref{fig:generation}, we empirically find that even leading image generation models, like DALL-E 3 and FLUX, struggle to generate content that conforms to given object orientation or spatial relationship conditions. Our model can help distinguish whether the generated image follows the given spatial condition, demonstrating its potential as a reward model to guide generative models in adhering to the desired orientation- and perspective-based spatial concepts.

\subsection{3D Models Orientation Voting} Many existing 3D data exhibit varied orientations, with some even tilted relative to the coordinate axes. As shown in Fig.~\ref{fig:case}, our method achieves consistent orientation predictions across multi-view images, enabling robust voting for 3D object's orientation. Accurately estimating the orientation of 3D models is valuable for further scaling up rendering images with orientation labels or adjusting the poses of 3D objects to a desired direction.

\section{Conclusion}
In this paper, we present Orient Anything, a practical approach for estimating object orientation from single images. We design an automatic and reliable 3D object annotation and rendering pipeline, allowing us to collect large-scale images with precise orientation annotations.
To fully exploit the value of the new dataset, we design an orientation probability distribution fitting task for robust orientation estimation, and improve synthetic-to-real transfer performance by incorporating real-world knowledge and reducing the domain gap. As a result, Orient Anything achieves impressive zero-shot object orientation estimation in real-world images and can serve as a foundational tool for enabling applications like complex spatial understanding and generation scoring.

\vspace{5\baselineskip}

{
    \small
    \bibliographystyle{ieeenat_fullname}
    \bibliography{main}

\begin{thebibliography}{50}
\providecommand{\natexlab}[1]{#1}
\providecommand{\url}[1]{\texttt{#1}}
\expandafter\ifx\csname urlstyle\endcsname\relax
  \providecommand{\doi}[1]{doi: #1}\else
  \providecommand{\doi}{doi: \begingroup \urlstyle{rm}\Url}\fi

\bibitem[Achlioptas et~al.(2020)Achlioptas, Abdelreheem, Xia, Elhoseiny, and Guibas]{achlioptas2020referit3d}
Panos Achlioptas, Ahmed Abdelreheem, Fei Xia, Mohamed Elhoseiny, and Leonidas Guibas.
\newblock Referit3d: Neural listeners for fine-grained 3d object identification in real-world scenes.
\newblock In \emph{Computer Vision--ECCV 2020: 16th European Conference, Glasgow, UK, August 23--28, 2020, Proceedings, Part I 16}, pages 422--440. Springer, 2020.

\bibitem[Ahmadyan et~al.(2021)Ahmadyan, Zhang, Ablavatski, Wei, and Grundmann]{ahmadyan2021objectron}
Adel Ahmadyan, Liangkai Zhang, Artsiom Ablavatski, Jianing Wei, and Matthias Grundmann.
\newblock Objectron: A large scale dataset of object-centric videos in the wild with pose annotations.
\newblock In \emph{Proceedings of the IEEE/CVF conference on computer vision and pattern recognition}, pages 7822--7831, 2021.

\bibitem[Azuma et~al.(2022)Azuma, Miyanishi, Kurita, and Kawanabe]{azuma2022scanqa}
Daichi Azuma, Taiki Miyanishi, Shuhei Kurita, and Motoaki Kawanabe.
\newblock Scanqa: 3d question answering for spatial scene understanding.
\newblock In \emph{proceedings of the IEEE/CVF conference on computer vision and pattern recognition}, pages 19129--19139, 2022.

\bibitem[Baruch et~al.(2021)Baruch, Chen, Dehghan, Dimry, Feigin, Fu, Gebauer, Joffe, Kurz, Schwartz, et~al.]{baruch2021arkitscenes}
Gilad Baruch, Zhuoyuan Chen, Afshin Dehghan, Tal Dimry, Yuri Feigin, Peter Fu, Thomas Gebauer, Brandon Joffe, Daniel Kurz, Arik Schwartz, et~al.
\newblock Arkitscenes: A diverse real-world dataset for 3d indoor scene understanding using mobile rgb-d data.
\newblock \emph{arXiv preprint arXiv:2111.08897}, 2021.

\bibitem[Betker et~al.(2023)Betker, Goh, Jing, Brooks, Wang, Li, Ouyang, Zhuang, Lee, Guo, et~al.]{betker2023improving}
James Betker, Gabriel Goh, Li Jing, Tim Brooks, Jianfeng Wang, Linjie Li, Long Ouyang, Juntang Zhuang, Joyce Lee, Yufei Guo, et~al.
\newblock Improving image generation with better captions.
\newblock \emph{Computer Science. https://cdn. openai. com/papers/dall-e-3. pdf}, 2\penalty0 (3):\penalty0 8, 2023.

\bibitem[Brazil et~al.(2023)Brazil, Kumar, Straub, Ravi, Johnson, and Gkioxari]{brazil2023omni3d}
Garrick Brazil, Abhinav Kumar, Julian Straub, Nikhila Ravi, Justin Johnson, and Georgia Gkioxari.
\newblock Omni3d: A large benchmark and model for 3d object detection in the wild.
\newblock In \emph{Proceedings of the IEEE/CVF conference on computer vision and pattern recognition}, pages 13154--13164, 2023.

\bibitem[Caesar et~al.(2020)Caesar, Bankiti, Lang, Vora, Liong, Xu, Krishnan, Pan, Baldan, and Beijbom]{caesar2020nuscenes}
Holger Caesar, Varun Bankiti, Alex~H Lang, Sourabh Vora, Venice~Erin Liong, Qiang Xu, Anush Krishnan, Yu Pan, Giancarlo Baldan, and Oscar Beijbom.
\newblock nuscenes: A multimodal dataset for autonomous driving.
\newblock In \emph{Proceedings of the IEEE/CVF conference on computer vision and pattern recognition}, pages 11621--11631, 2020.

\bibitem[Chen et~al.(2020)Chen, Chang, and Nie{\ss}ner]{chen2020scanrefer}
Dave~Zhenyu Chen, Angel~X Chang, and Matthias Nie{\ss}ner.
\newblock Scanrefer: 3d object localization in rgb-d scans using natural language.
\newblock In \emph{European conference on computer vision}, pages 202--221. Springer, 2020.

\bibitem[Deitke et~al.(2023)Deitke, Schwenk, Salvador, Weihs, Michel, VanderBilt, Schmidt, Ehsani, Kembhavi, and Farhadi]{deitke2023objaverse}
Matt Deitke, Dustin Schwenk, Jordi Salvador, Luca Weihs, Oscar Michel, Eli VanderBilt, Ludwig Schmidt, Kiana Ehsani, Aniruddha Kembhavi, and Ali Farhadi.
\newblock Objaverse: A universe of annotated 3d objects.
\newblock In \emph{Proceedings of the IEEE/CVF Conference on Computer Vision and Pattern Recognition}, pages 13142--13153, 2023.

\bibitem[Fan et~al.(2024)Fan, Pan, Wang, Jiang, Xu, and Wang]{fan2024pope}
Zhiwen Fan, Panwang Pan, Peihao Wang, Yifan Jiang, Dejia Xu, and Zhangyang Wang.
\newblock Pope: 6-dof promptable pose estimation of any object in any scene with one reference.
\newblock In \emph{Proceedings of the IEEE/CVF Conference on Computer Vision and Pattern Recognition}, pages 7771--7781, 2024.

\bibitem[Geiger et~al.(2012)Geiger, Lenz, and Urtasun]{geiger2012kitti}
Andreas Geiger, Philip Lenz, and Raquel Urtasun.
\newblock Are we ready for autonomous driving? the kitti vision benchmark suite.
\newblock In \emph{2012 IEEE conference on computer vision and pattern recognition}, pages 3354--3361. IEEE, 2012.

\bibitem[Goodwin et~al.(2022)Goodwin, Vaze, Havoutis, and Posner]{goodwin2022zero}
Walter Goodwin, Sagar Vaze, Ioannis Havoutis, and Ingmar Posner.
\newblock Zero-shot category-level object pose estimation.
\newblock In \emph{European Conference on Computer Vision}, pages 516--532. Springer, 2022.

\bibitem[G{\'o}ral et~al.(2024)G{\'o}ral, Ziarko, Nauman, and Wo{\l}czyk]{goral2024seeing}
Gracjan G{\'o}ral, Alicja Ziarko, Michal Nauman, and Maciej Wo{\l}czyk.
\newblock Seeing through their eyes: Evaluating visual perspective taking in vision language models.
\newblock \emph{arXiv preprint arXiv:2409.12969}, 2024.

\bibitem[Guan et~al.(2024)Guan, Hao, Wu, Li, and Fang]{guan2024survey}
Jian Guan, Yingming Hao, Qingxiao Wu, Sicong Li, and Yingjian Fang.
\newblock A survey of 6dof object pose estimation methods for different application scenarios.
\newblock \emph{Sensors}, 24\penalty0 (4):\penalty0 1076, 2024.

\bibitem[Han et~al.(2021)Han, Ding, Li, and Xia]{han2021align}
Jiaming Han, Jian Ding, Jie Li, and Gui-Song Xia.
\newblock Align deep features for oriented object detection.
\newblock \emph{IEEE transactions on geoscience and remote sensing}, 60:\penalty0 1--11, 2021.

\bibitem[He et~al.(2022)He, Chen, Xie, Li, Doll{\'a}r, and Girshick]{he2022masked}
Kaiming He, Xinlei Chen, Saining Xie, Yanghao Li, Piotr Doll{\'a}r, and Ross Girshick.
\newblock Masked autoencoders are scalable vision learners.
\newblock In \emph{Proceedings of the IEEE/CVF conference on computer vision and pattern recognition}, pages 16000--16009, 2022.

\bibitem[Huang et~al.(2024)Huang, Li, Chen, Ren, Lin, Morstatter, and Xu]{huang2024orientdream}
Yuzhong Huang, Zhong Li, Zhang Chen, Zhiyuan Ren, Guosheng Lin, Fred Morstatter, and Yi Xu.
\newblock Orientdream: Streamlining text-to-3d generation with explicit orientation control.
\newblock \emph{arXiv preprint arXiv:2406.10000}, 2024.

\bibitem[Hurst et~al.(2024)Hurst, Lerer, Goucher, Perelman, Ramesh, Clark, Ostrow, Welihinda, Hayes, Radford, et~al.]{hurst2024gpt}
Aaron Hurst, Adam Lerer, Adam~P Goucher, Adam Perelman, Aditya Ramesh, Aidan Clark, AJ Ostrow, Akila Welihinda, Alan Hayes, Alec Radford, et~al.
\newblock Gpt-4o system card.
\newblock \emph{arXiv preprint arXiv:2410.21276}, 2024.

\bibitem[Ke et~al.(2024)Ke, Obukhov, Huang, Metzger, Daudt, and Schindler]{ke2024repurposing}
Bingxin Ke, Anton Obukhov, Shengyu Huang, Nando Metzger, Rodrigo~Caye Daudt, and Konrad Schindler.
\newblock Repurposing diffusion-based image generators for monocular depth estimation.
\newblock In \emph{Proceedings of the IEEE/CVF Conference on Computer Vision and Pattern Recognition}, pages 9492--9502, 2024.

\bibitem[Kirillov et~al.(2023)Kirillov, Mintun, Ravi, Mao, Rolland, Gustafson, Xiao, Whitehead, Berg, Lo, et~al.]{kirillov2023segment}
Alexander Kirillov, Eric Mintun, Nikhila Ravi, Hanzi Mao, Chloe Rolland, Laura Gustafson, Tete Xiao, Spencer Whitehead, Alexander~C Berg, Wan-Yen Lo, et~al.
\newblock Segment anything.
\newblock In \emph{Proceedings of the IEEE/CVF International Conference on Computer Vision}, pages 4015--4026, 2023.

\bibitem[Labs(2024)]{flux}
Black~Forest Labs.
\newblock Flux, 2024.

\bibitem[Li et~al.(2022)Li, Zhang, Zhang, Yang, Li, Zhong, Wang, Yuan, Zhang, Hwang, et~al.]{li2022grounded}
Liunian~Harold Li, Pengchuan Zhang, Haotian Zhang, Jianwei Yang, Chunyuan Li, Yiwu Zhong, Lijuan Wang, Lu Yuan, Lei Zhang, Jenq-Neng Hwang, et~al.
\newblock Grounded language-image pre-training.
\newblock In \emph{Proceedings of the IEEE/CVF Conference on Computer Vision and Pattern Recognition}, pages 10965--10975, 2022.

\bibitem[Lin et~al.(2014)Lin, Maire, Belongie, Hays, Perona, Ramanan, Doll{\'a}r, and Zitnick]{lin2014microsoft}
Tsung-Yi Lin, Michael Maire, Serge Belongie, James Hays, Pietro Perona, Deva Ramanan, Piotr Doll{\'a}r, and C~Lawrence Zitnick.
\newblock Microsoft coco: Common objects in context.
\newblock In \emph{Computer Vision--ECCV 2014: 13th European Conference, Zurich, Switzerland, September 6-12, 2014, Proceedings, Part V 13}, pages 740--755. Springer, 2014.

\bibitem[Liu et~al.(2024)Liu, Li, Wu, and Lee]{liu2024visual}
Haotian Liu, Chunyuan Li, Qingyang Wu, and Yong~Jae Lee.
\newblock Visual instruction tuning.
\newblock \emph{Advances in neural information processing systems}, 36, 2024.

\bibitem[Liu et~al.(2023)Liu, Zeng, Ren, Li, Zhang, Yang, Jiang, Li, Yang, Su, et~al.]{liu2023grounding}
Shilong Liu, Zhaoyang Zeng, Tianhe Ren, Feng Li, Hao Zhang, Jie Yang, Qing Jiang, Chunyuan Li, Jianwei Yang, Hang Su, et~al.
\newblock Grounding dino: Marrying dino with grounded pre-training for open-set object detection.
\newblock \emph{arXiv preprint arXiv:2303.05499}, 2023.

\bibitem[Loshchilov(2017)]{loshchilov2017decoupled}
I Loshchilov.
\newblock Decoupled weight decay regularization.
\newblock \emph{arXiv preprint arXiv:1711.05101}, 2017.

\bibitem[Lowe(2004)]{lowe2004distinctive}
David~G Lowe.
\newblock Distinctive image features from scale-invariant keypoints.
\newblock \emph{International journal of computer vision}, 60:\penalty0 91--110, 2004.

\bibitem[Ma et~al.(2022)Ma, Yong, Zheng, Li, Liang, Zhu, and Huang]{ma2022sqa3d}
Xiaojian Ma, Silong Yong, Zilong Zheng, Qing Li, Yitao Liang, Song-Chun Zhu, and Siyuan Huang.
\newblock Sqa3d: Situated question answering in 3d scenes.
\newblock \emph{arXiv preprint arXiv:2210.07474}, 2022.

\bibitem[Nguyen et~al.(2024)Nguyen, Groueix, Ponimatkin, Hu, Marlet, Salzmann, and Lepetit]{nguyen2024nope}
Van~Nguyen Nguyen, Thibault Groueix, Georgy Ponimatkin, Yinlin Hu, Renaud Marlet, Mathieu Salzmann, and Vincent Lepetit.
\newblock Nope: Novel object pose estimation from a single image.
\newblock In \emph{Proceedings of the IEEE/CVF Conference on Computer Vision and Pattern Recognition}, pages 17923--17932, 2024.

\bibitem[Oquab et~al.(2023)Oquab, Darcet, Moutakanni, Vo, Szafraniec, Khalidov, Fernandez, Haziza, Massa, El-Nouby, et~al.]{oquab2023dinov2}
Maxime Oquab, Timoth{\'e}e Darcet, Th{\'e}o Moutakanni, Huy Vo, Marc Szafraniec, Vasil Khalidov, Pierre Fernandez, Daniel Haziza, Francisco Massa, Alaaeldin El-Nouby, et~al.
\newblock Dinov2: Learning robust visual features without supervision.
\newblock \emph{arXiv preprint arXiv:2304.07193}, 2023.

\bibitem[Peng et~al.(2019)Peng, Liu, Huang, Zhou, and Bao]{peng2019pvnet}
Sida Peng, Yuan Liu, Qixing Huang, Xiaowei Zhou, and Hujun Bao.
\newblock Pvnet: Pixel-wise voting network for 6dof pose estimation.
\newblock In \emph{Proceedings of the IEEE/CVF conference on computer vision and pattern recognition}, pages 4561--4570, 2019.

\bibitem[Radford et~al.(2021)Radford, Kim, Hallacy, Ramesh, Goh, Agarwal, Sastry, Askell, Mishkin, Clark, et~al.]{radford2021learning}
Alec Radford, Jong~Wook Kim, Chris Hallacy, Aditya Ramesh, Gabriel Goh, Sandhini Agarwal, Girish Sastry, Amanda Askell, Pamela Mishkin, Jack Clark, et~al.
\newblock Learning transferable visual models from natural language supervision.
\newblock In \emph{International conference on machine learning}, pages 8748--8763. PMLR, 2021.

\bibitem[Raji{\v{c}} et~al.(2023)Raji{\v{c}}, Ke, Tai, Tang, Danelljan, and Yu]{rajivc2023segment}
Frano Raji{\v{c}}, Lei Ke, Yu-Wing Tai, Chi-Keung Tang, Martin Danelljan, and Fisher Yu.
\newblock Segment anything meets point tracking.
\newblock \emph{arXiv preprint arXiv:2307.01197}, 2023.

\bibitem[Ravi et~al.(2024)Ravi, Gabeur, Hu, Hu, Ryali, Ma, Khedr, R{\"a}dle, Rolland, Gustafson, et~al.]{ravi2024sam}
Nikhila Ravi, Valentin Gabeur, Yuan-Ting Hu, Ronghang Hu, Chaitanya Ryali, Tengyu Ma, Haitham Khedr, Roman R{\"a}dle, Chloe Rolland, Laura Gustafson, et~al.
\newblock Sam 2: Segment anything in images and videos.
\newblock \emph{arXiv preprint arXiv:2408.00714}, 2024.

\bibitem[Ren et~al.(2024)Ren, Liu, Zeng, Lin, Li, Cao, Chen, Huang, Chen, Yan, Zeng, Zhang, Li, Yang, Li, Jiang, and Zhang]{ren2024grounded}
Tianhe Ren, Shilong Liu, Ailing Zeng, Jing Lin, Kunchang Li, He Cao, Jiayu Chen, Xinyu Huang, Yukang Chen, Feng Yan, Zhaoyang Zeng, Hao Zhang, Feng Li, Jie Yang, Hongyang Li, Qing Jiang, and Lei Zhang.
\newblock Grounded sam: Assembling open-world models for diverse visual tasks, 2024.

\bibitem[Schuhmann et~al.(2021)Schuhmann, Vencu, Beaumont, Kaczmarczyk, Mullis, Katta, Coombes, Jitsev, and Komatsuzaki]{schuhmann2021laion}
Christoph Schuhmann, Richard Vencu, Romain Beaumont, Robert Kaczmarczyk, Clayton Mullis, Aarush Katta, Theo Coombes, Jenia Jitsev, and Aran Komatsuzaki.
\newblock Laion-400m: Open dataset of clip-filtered 400 million image-text pairs.
\newblock \emph{arXiv preprint arXiv:2111.02114}, 2021.

\bibitem[Schuhmann et~al.(2022)Schuhmann, Beaumont, Vencu, Gordon, Wightman, Cherti, Coombes, Katta, Mullis, Wortsman, et~al.]{schuhmann2022laion}
Christoph Schuhmann, Romain Beaumont, Richard Vencu, Cade Gordon, Ross Wightman, Mehdi Cherti, Theo Coombes, Aarush Katta, Clayton Mullis, Mitchell Wortsman, et~al.
\newblock Laion-5b: An open large-scale dataset for training next generation image-text models.
\newblock \emph{Advances in Neural Information Processing Systems}, 35:\penalty0 25278--25294, 2022.

\bibitem[Shi et~al.(2023)Shi, Wang, Ye, Long, Li, and Yang]{shi2023mvdream}
Yichun Shi, Peng Wang, Jianglong Ye, Mai Long, Kejie Li, and Xiao Yang.
\newblock Mvdream: Multi-view diffusion for 3d generation.
\newblock \emph{arXiv preprint arXiv:2308.16512}, 2023.

\bibitem[Song et~al.(2015)Song, Lichtenberg, and Xiao]{song2015sun}
Shuran Song, Samuel~P Lichtenberg, and Jianxiong Xiao.
\newblock Sun rgb-d: A rgb-d scene understanding benchmark suite.
\newblock In \emph{Proceedings of the IEEE conference on computer vision and pattern recognition}, pages 567--576, 2015.

\bibitem[Sundermeyer et~al.(2018)Sundermeyer, Marton, Durner, Brucker, and Triebel]{sundermeyer2018implicit}
Martin Sundermeyer, Zoltan-Csaba Marton, Maximilian Durner, Manuel Brucker, and Rudolph Triebel.
\newblock Implicit 3d orientation learning for 6d object detection from rgb images.
\newblock In \emph{Proceedings of the european conference on computer vision (ECCV)}, pages 699--715, 2018.

\bibitem[Team et~al.(2023)Team, Anil, Borgeaud, Alayrac, Yu, Soricut, Schalkwyk, Dai, Hauth, Millican, et~al.]{team2023gemini}
Gemini Team, Rohan Anil, Sebastian Borgeaud, Jean-Baptiste Alayrac, Jiahui Yu, Radu Soricut, Johan Schalkwyk, Andrew~M Dai, Anja Hauth, Katie Millican, et~al.
\newblock Gemini: a family of highly capable multimodal models.
\newblock \emph{arXiv preprint arXiv:2312.11805}, 2023.

\bibitem[Team et~al.(2024)Team, Georgiev, Lei, Burnell, Bai, Gulati, Tanzer, Vincent, Pan, Wang, et~al.]{team2024gemini}
Gemini Team, Petko Georgiev, Ving~Ian Lei, Ryan Burnell, Libin Bai, Anmol Gulati, Garrett Tanzer, Damien Vincent, Zhufeng Pan, Shibo Wang, et~al.
\newblock Gemini 1.5: Unlocking multimodal understanding across millions of tokens of context.
\newblock \emph{arXiv preprint arXiv:2403.05530}, 2024.

\bibitem[Wang et~al.(2023)Wang, Chang, Cai, Li, Hariharan, Holynski, and Snavely]{wang2023tracking}
Qianqian Wang, Yen-Yu Chang, Ruojin Cai, Zhengqi Li, Bharath Hariharan, Aleksander Holynski, and Noah Snavely.
\newblock Tracking everything everywhere all at once.
\newblock In \emph{Proceedings of the IEEE/CVF International Conference on Computer Vision}, pages 19795--19806, 2023.

\bibitem[Wang et~al.(2024)Wang, Mao, Zhu, Xu, Lyu, Li, Chen, Zhang, Chen, Xue, et~al.]{wang2024embodiedscan}
Tai Wang, Xiaohan Mao, Chenming Zhu, Runsen Xu, Ruiyuan Lyu, Peisen Li, Xiao Chen, Wenwei Zhang, Kai Chen, Tianfan Xue, et~al.
\newblock Embodiedscan: A holistic multi-modal 3d perception suite towards embodied ai.
\newblock In \emph{Proceedings of the IEEE/CVF Conference on Computer Vision and Pattern Recognition}, pages 19757--19767, 2024.

\bibitem[Wei et~al.(2023)Wei, Ding, Park, Sajnani, Poulenard, Sridhar, and Guibas]{wei2023lego}
Qiuhong~Anna Wei, Sijie Ding, Jeong~Joon Park, Rahul Sajnani, Adrien Poulenard, Srinath Sridhar, and Leonidas Guibas.
\newblock Lego-net: Learning regular rearrangements of objects in rooms.
\newblock In \emph{Proceedings of the IEEE/CVF Conference on Computer Vision and Pattern Recognition}, pages 19037--19047, 2023.

\bibitem[Wu et~al.(2024)Wu, Li, and Liu]{wu2024human}
Zhen Wu, Jiaman Li, and C~Karen Liu.
\newblock Human-object interaction from human-level instructions.
\newblock \emph{arXiv preprint arXiv:2406.17840}, 2024.

\bibitem[Xie et~al.(2021)Xie, Cheng, Wang, Yao, and Han]{xie2021oriented}
Xingxing Xie, Gong Cheng, Jiabao Wang, Xiwen Yao, and Junwei Han.
\newblock Oriented r-cnn for object detection.
\newblock In \emph{Proceedings of the IEEE/CVF international conference on computer vision}, pages 3520--3529, 2021.

\bibitem[Yang et~al.(2024)Yang, Kang, Huang, Zhao, Xu, Feng, and Zhao]{yang2024depth}
Lihe Yang, Bingyi Kang, Zilong Huang, Zhen Zhao, Xiaogang Xu, Jiashi Feng, and Hengshuang Zhao.
\newblock Depth anything v2.
\newblock \emph{arXiv preprint arXiv:2406.09414}, 2024.

\bibitem[Zhang et~al.(2024)Zhang, Huang, Ma, Li, Luo, Xie, Qin, Luo, Li, Liu, et~al.]{zhang2024recognize}
Youcai Zhang, Xinyu Huang, Jinyu Ma, Zhaoyang Li, Zhaochuan Luo, Yanchun Xie, Yuzhuo Qin, Tong Luo, Yaqian Li, Shilong Liu, et~al.
\newblock Recognize anything: A strong image tagging model.
\newblock In \emph{Proceedings of the IEEE/CVF Conference on Computer Vision and Pattern Recognition}, pages 1724--1732, 2024.

\bibitem[Zhou et~al.(2017)Zhou, Zhao, Puig, Fidler, Barriuso, and Torralba]{zhou2017scene}
Bolei Zhou, Hang Zhao, Xavier Puig, Sanja Fidler, Adela Barriuso, and Antonio Torralba.
\newblock Scene parsing through ade20k dataset.
\newblock In \emph{Proceedings of the IEEE conference on computer vision and pattern recognition}, pages 633--641, 2017.

\end{thebibliography}
}

\newpage

\appendix

\begin{table}[t]
\setlength\tabcolsep{4pt}
\resizebox{\linewidth}{!}{
\begin{tabular}{cccccc}
\toprule
Category & Cube RCNN & Gemini & GPT-4o & Orient Anything (ViT-L) &  \\ \midrule
bed & 75\% & 15\% & 40\% & 100\%\textcolor{green}{(+25\%)} \\
monitor & 35\% & 50\% & 50\% & 100\%\textcolor{green}{(+50\%)} \\
oven & 50\% & 10\% & 65\% & 100\%\textcolor{green}{(+35\%)} \\
teddy bear & 20\% & 40\% & 45\% & 100\%\textcolor{green}{(+55\%)} \\
motorbike & 5\% & 20\% & 40\% & 95\%\textcolor{green}{(+55\%)} \\
parking meter & 40\% & 55\% & 65\% & 95\%\textcolor{green}{(+30\%)} \\
laptop & 65\% & 45\% & 50\% & 95\%\textcolor{green}{(+30\%)} \\
sheep & 15\% & 45\% & 45\% & 90\%\textcolor{green}{(+45\%)} \\
elephant & 5\% & 30\% & 55\% & 90\%\textcolor{green}{(+35\%)} \\
sofa & 5\% & 25\% & 50\% & 90\%\textcolor{green}{(+40\%)} \\
toilet & 55\% & 20\% & 50\% & 90\%\textcolor{green}{(+35\%)} \\
cell phone & 35\% & 75\% & 80\% & 90\%\textcolor{green}{(+10\%)} \\
microwave & 35\% & 25\% & 50\% & 90\%\textcolor{green}{(+40\%)} \\
clock & 20\% & 45\% & 60\% & 90\%\textcolor{green}{(+30\%)} \\
bus & 10\% & 20\% & 40\% & 85\%\textcolor{green}{(+45\%)} \\
traffic light & 0\% & 35\% & 50\% & 85\%\textcolor{green}{(+35\%)} \\
stop sign & 0\% & 70\% & 75\% & 85\%\textcolor{green}{(+10\%)} \\
bench & 20\% & 20\% & 20\% & 85\%\textcolor{green}{(+65\%)} \\
bear & 5\% & 30\% & 40\% & 85\%\textcolor{green}{(+45\%)} \\
zebra & 5\% & 30\% & 50\% & 85\%\textcolor{green}{(+35\%)} \\
sink & 0\% & 0\% & 30\% & 85\%\textcolor{green}{(+55\%)} \\
cat & 20\% & 45\% & 60\% & 80\%\textcolor{green}{(+20\%)}\\
dog & 10\% & 35\% & 60\% & 80\%\textcolor{green}{(+20\%)} \\
horse & 10\% & 50\% & 35\% & 80\%\textcolor{green}{(+30\%)} \\
chair & 10\% & 20\% & 30\% & 80\%\textcolor{green}{(+50\%)} \\
book & 45\% & 80\% & 45\% & 80\%\textcolor{green}{(+0\%)} \\
car & 10\% & 40\% & 45\% & 75\%\textcolor{green}{(+30\%)} \\
truck & 15\% & 35\% & 60\% & 75\%\textcolor{green}{(+15\%)} \\
cow & 20\% & 40\% & 40\% & 75\%\textcolor{green}{(+35\%)} \\
person & 5\% & 35\% & 40\% & 70\%\textcolor{green}{(+30\%)} \\
aeroplane & 15\% & 20\% & 60\% & 70\%\textcolor{green}{(+10\%)} \\
refrigerator & 20\% & 25\% & 55\% & 70\%\textcolor{green}{(+15\%)} \\
bird & 10\% & 25\% & 60\% & 65\%\textcolor{green}{(+5\%)} \\
giraffe & 15\% & 30\% & 55\% & 65\%\textcolor{green}{(+10\%)} \\
train & 30\% & 15\% & 60\% & 60\%\textcolor{green}{(+0\%)} \\
fire hydrant & 20\% & 20\% & 30\% & 55\%\textcolor{green}{(+25\%)} \\
boat & 10\% & 25\% & 45\% & 50\%\textcolor{green}{(+5\%)} \\
backpack & 40\% & 65\% & 50\% & 50\%\textcolor{red}{(-15\%)} \\
mouse & 15\% & 0\% & 0\% & 50\%\textcolor{green}{(+35\%)} \\
kite & 5\% & 40\% & 60\% & 45\%\textcolor{red}{(-15\%)} \\
hair drier & 0\% & 20\% & 55\% & 45\%\textcolor{red}{(-10\%)} \\
bicycle & 5\% & 30\% & 30\% & 40\%\textcolor{green}{(+10\%)} \\
toaster & 50\% & 25\% & 30\% & 40\%\textcolor{red}{(-10\%)} \\
remote & 10\% & 5\% & 10\% & 5\%\textcolor{red}{(-5\%)} \\
keyboard & 10\% & 0\% & 0\% & 0\%\textcolor{red}{(-10\%)} \\ \bottomrule
\end{tabular}}
\vspace{-0.6\baselineskip}
\caption{Detailed horizontal direction recognition accuracy for each object category in COCO that is annotated with front face and orientation. The differences between Orient Anything and the best results achieved by other alternative methods are also provided.}
\label{tab:coco}
\end{table}

\vspace{-2\baselineskip}
\section{Detailed Results on COCO Benchmark}
In Tab.~\ref{tab:coco}, we provide the detailed horizontal direction recognition accuracy for each object category in COCO that is annotated with front face and orientation.

Our model achieves excellent performance across most object categories with clear orientations, attaining an accuracy exceeding 80\%. However, it performs relatively poorly in categories where the distinction between front and back is ambiguous or the objects are too small. Compared to previous alternatives, Orient Anything achieves significantly better accuracy in most categories than the best results achieved by previous models.

\section{Visualization of Real-image Benchmarks}
\begin{figure}[t]
    \centering
    Cube RCNN \ \ \ \ \ \ \ \ \ \ \ \ \ \ \ \ \ \ \ \ \ \ \ \ \ \ \ \  Orient Anything \ \ \ \ \ \ \ \ \\
    \includegraphics[width=1\linewidth]{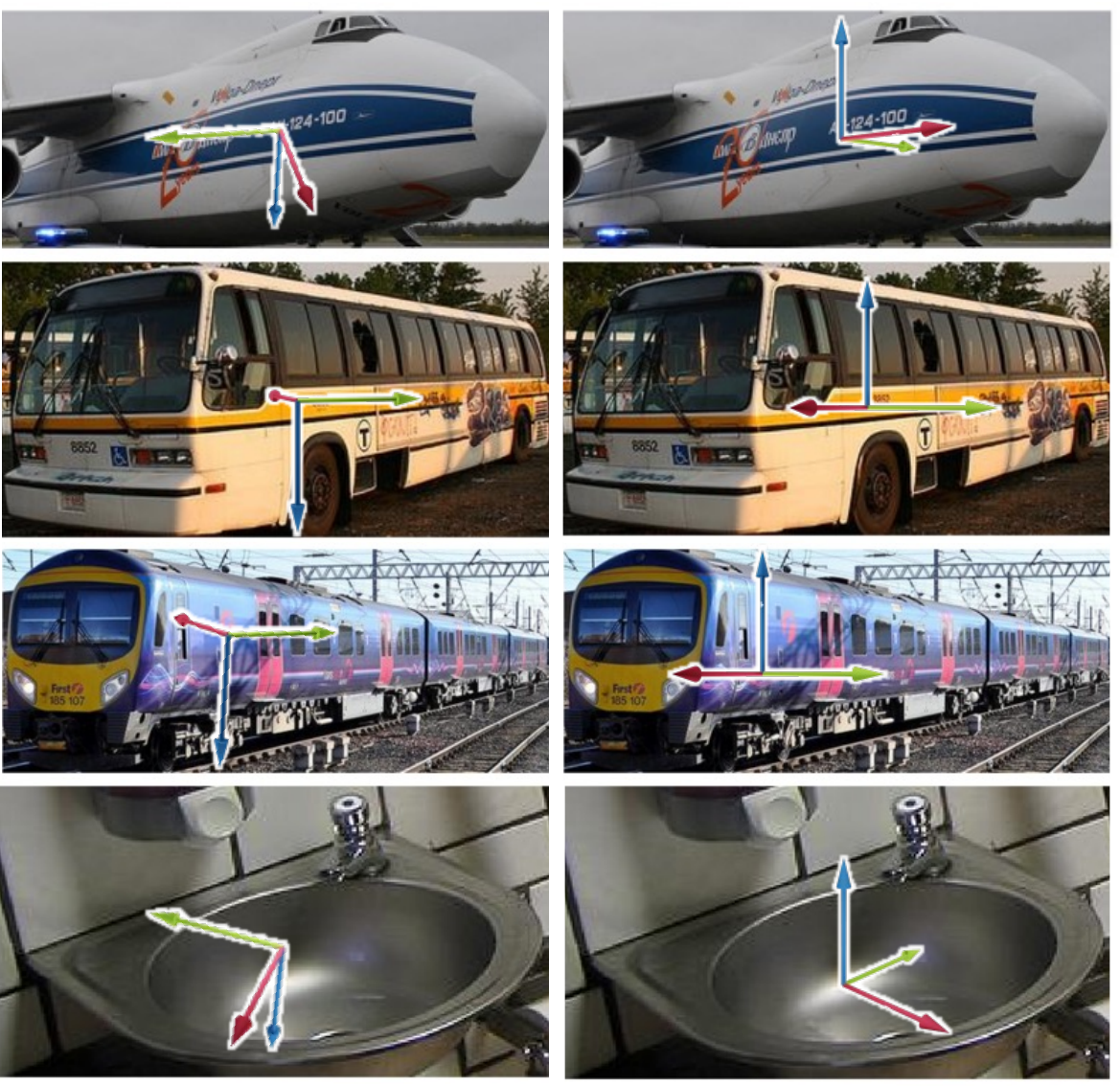}
\includegraphics[width=1\linewidth]{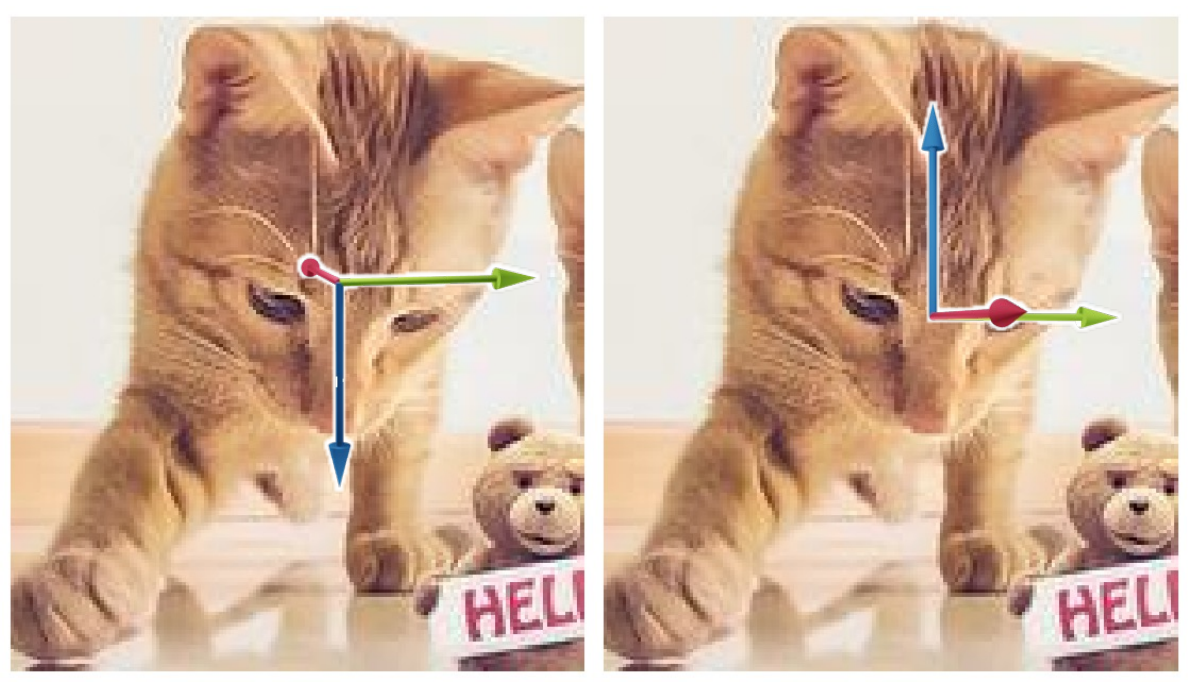}
\includegraphics[width=1\linewidth]{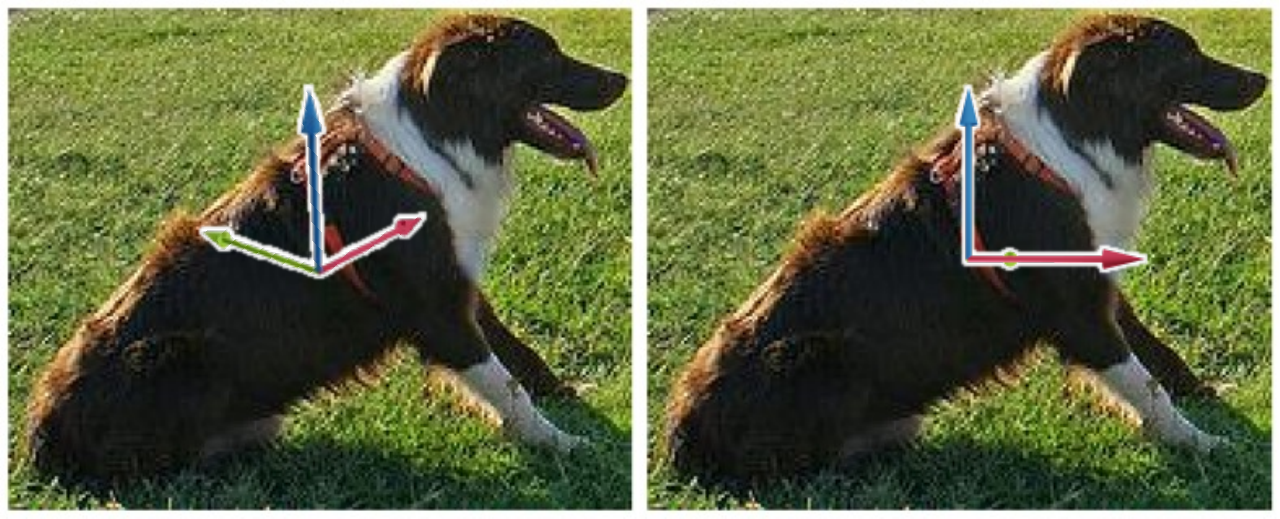}
\includegraphics[width=1\linewidth]{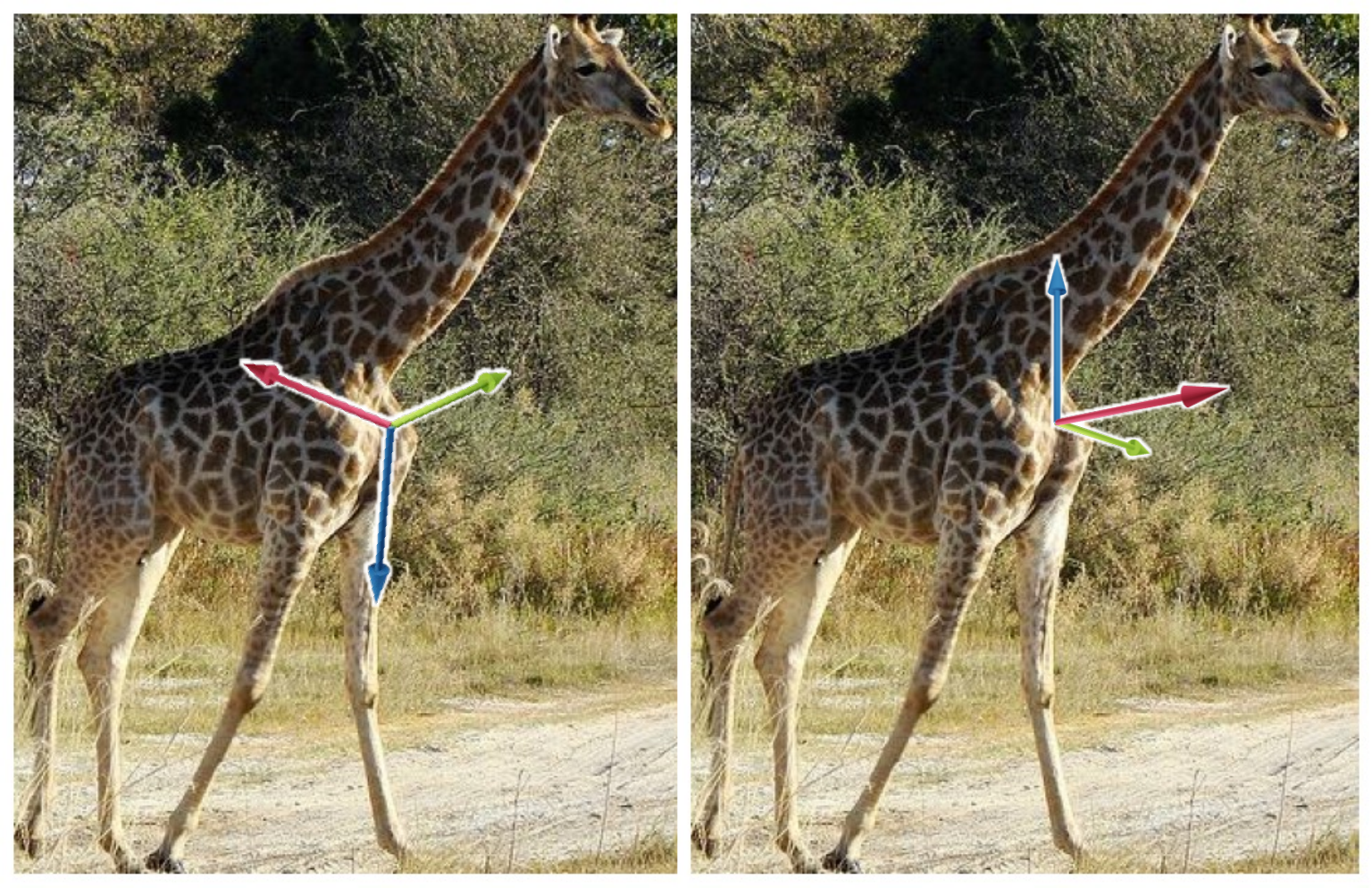}
\vspace{-1.5\baselineskip}
    \caption{Qualitative results on COCO}
    \label{fig:coco}
\end{figure}

\begin{figure}[t]
    \centering
    \includegraphics[width=1\linewidth]{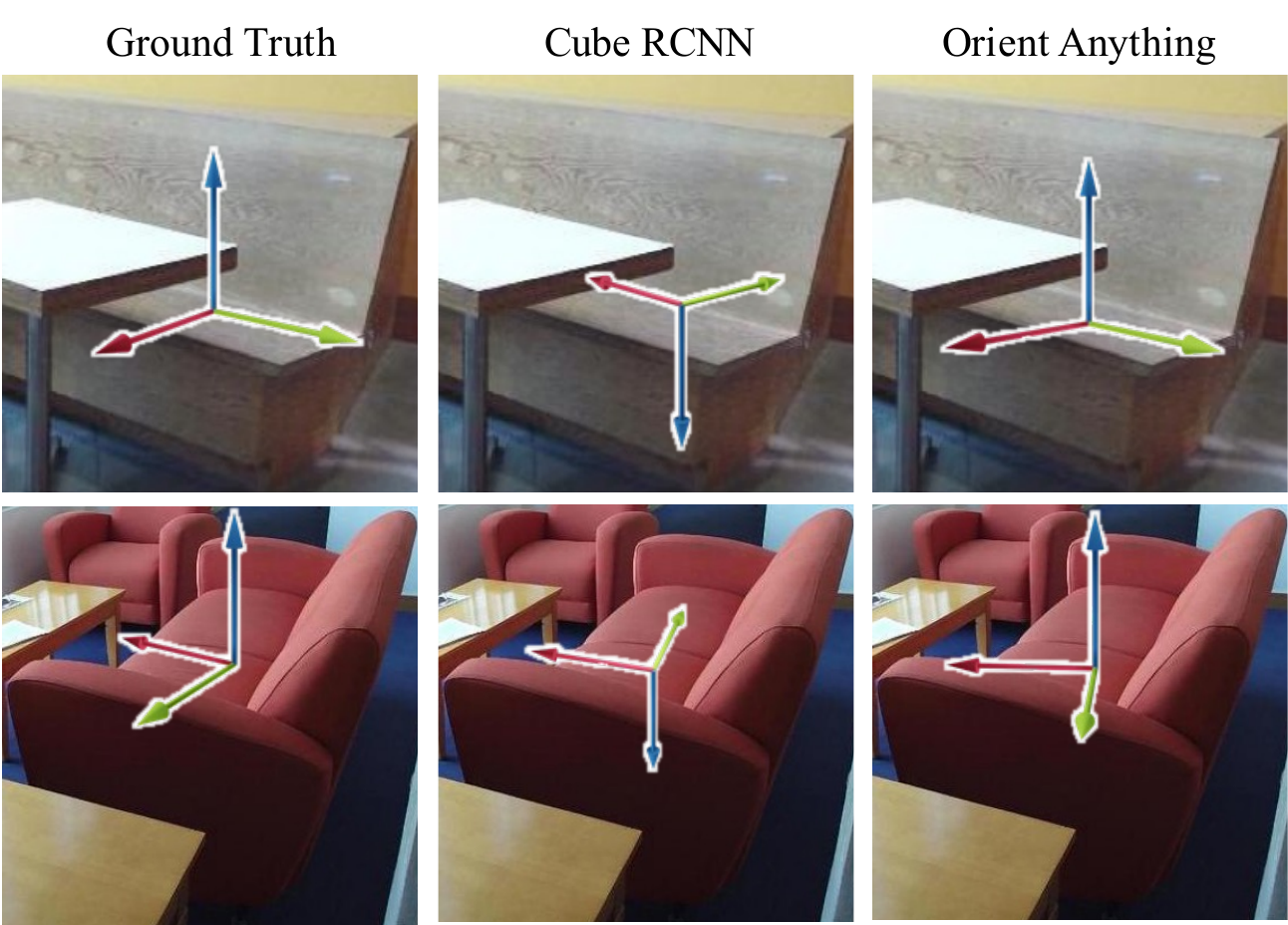}
\vspace{-1.5\baselineskip}
    \caption{Qualitative results on SUN RGB-D.}
    \label{fig:sunrgbd}
\end{figure}

\begin{figure}[t]
    \centering
    \includegraphics[width=1\linewidth]{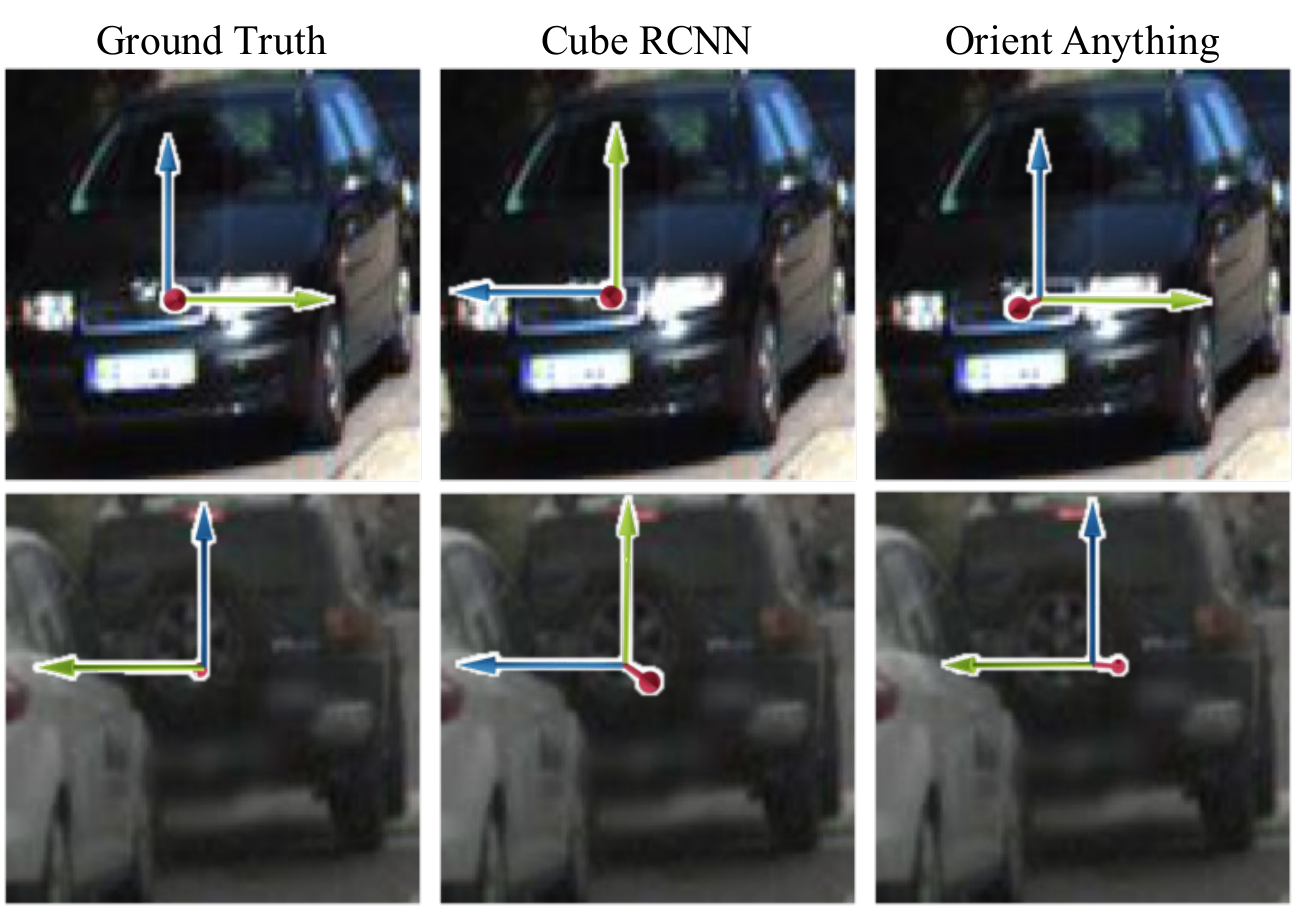}
\vspace{-1.5\baselineskip}
    \caption{Qualitative results on KITTI and nuScenes.}
    \label{fig:kitti}
\end{figure}

\begin{figure}[t]
    \centering
    \includegraphics[width=1\linewidth]{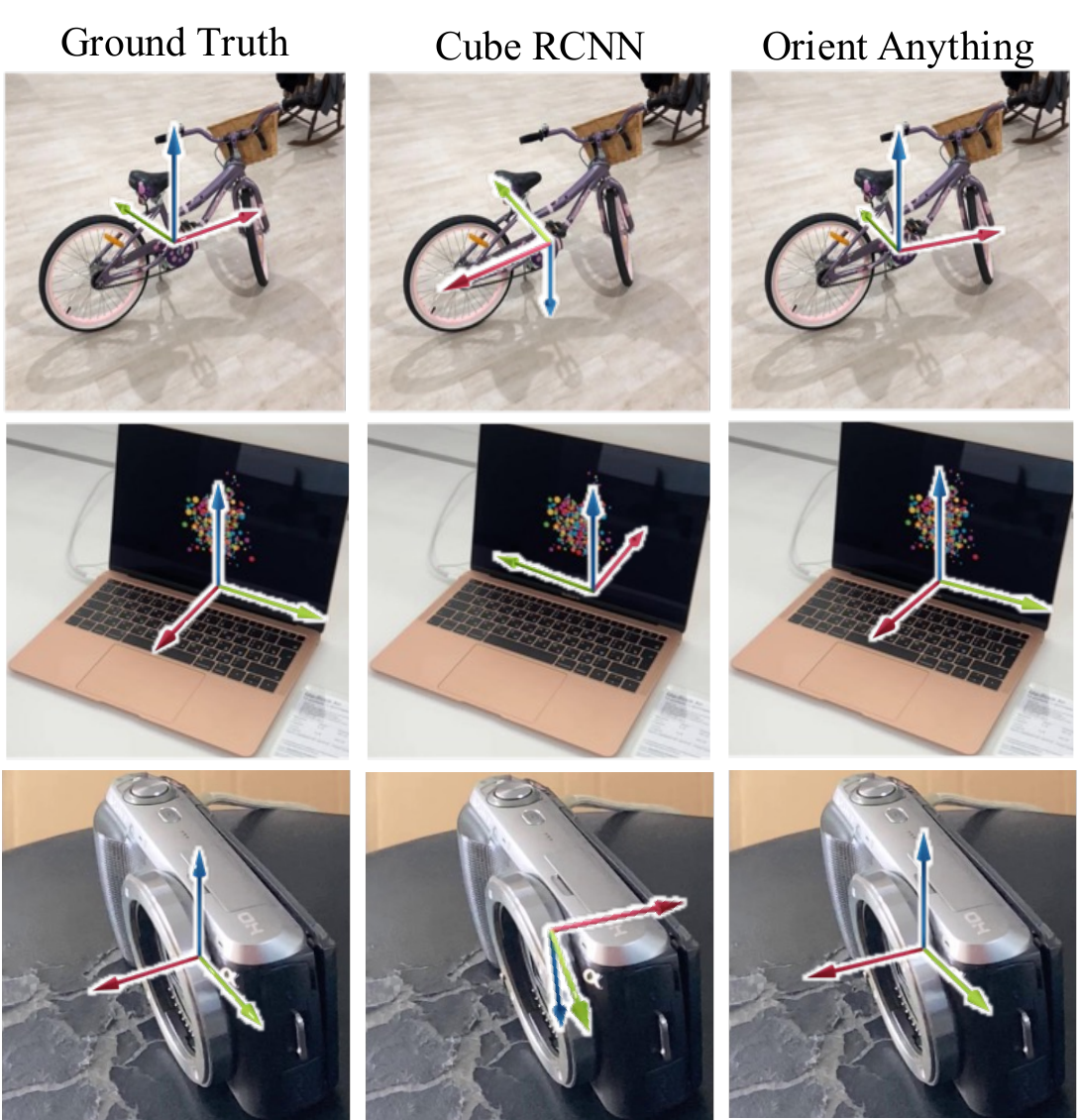}
\vspace{-1.5\baselineskip}
    \caption{Qualitative results on Objectron.}
    \label{fig:objectron}
\end{figure}

\begin{figure}[t]
    \centering
    \includegraphics[width=1\linewidth]{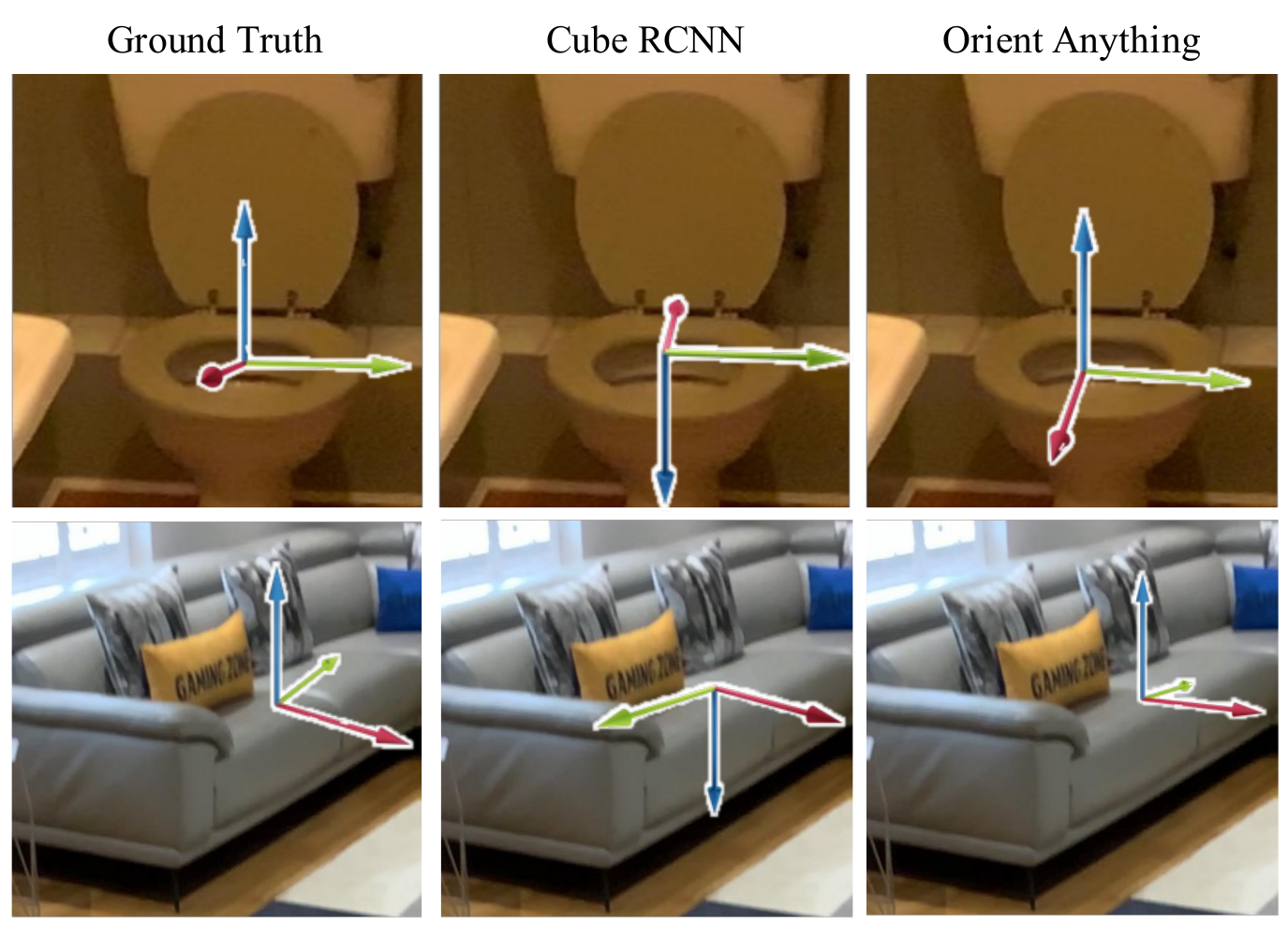}
\vspace{-1.5\baselineskip}
    \caption{Qualitative results on ARKitScenes.}
    \label{fig:ark}
\end{figure}

In Fig.~\ref{fig:coco}, \ref{fig:sunrgbd}, \ref{fig:kitti}, \ref{fig:objectron} and \ref{fig:ark}, we present the qualitative results on objects of COCO, SUN RGB-D, KITTI, nuScenes, Objectron, and ARKitScenes, respectively. Our model can robustly and accurately predict the object orientation in images of various sources and resolutions.

\section{More Visualizations of Images in The Wild}
\begin{figure}[t]
    \centering
    \includegraphics[width=1\linewidth]{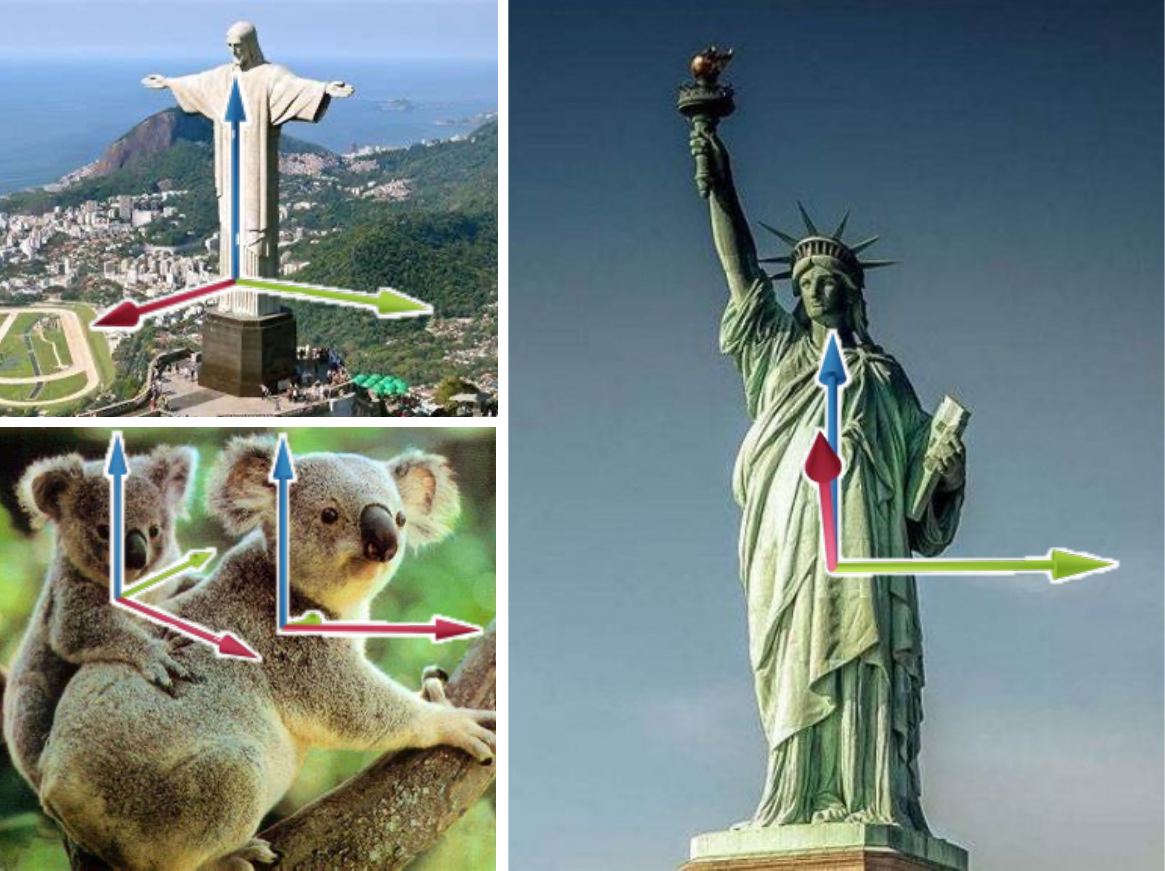}
    \includegraphics[width=0.99\linewidth]{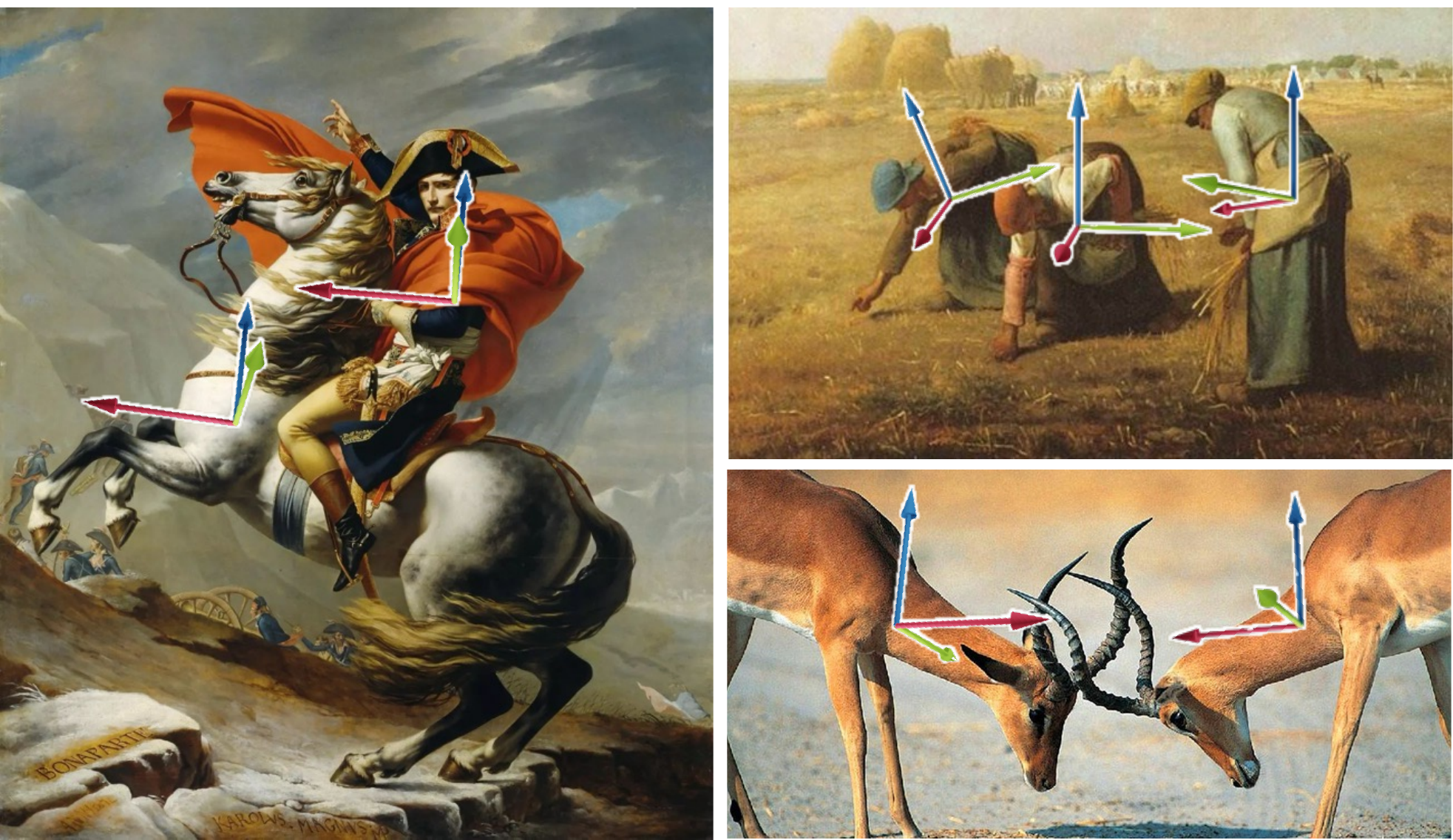}
    \vspace{-0.5\baselineskip}
    \caption{More visualization of images in the wild.}
    \vspace{-1\baselineskip}
    \label{fig:wild}
\end{figure}

In Fig.~\ref{fig:wild}, we present more visualizations of images from various domains containing different objects. In these images, our model shows consistently accurate orientation prediction results, further highlighting the impressive zero-shot capability of our Orient Anything.

\section{Visualization of Ori-Bench}
\begin{figure*}[t]
    \centering
    \vspace{-2\baselineskip}
    \includegraphics[width=0.7\linewidth]{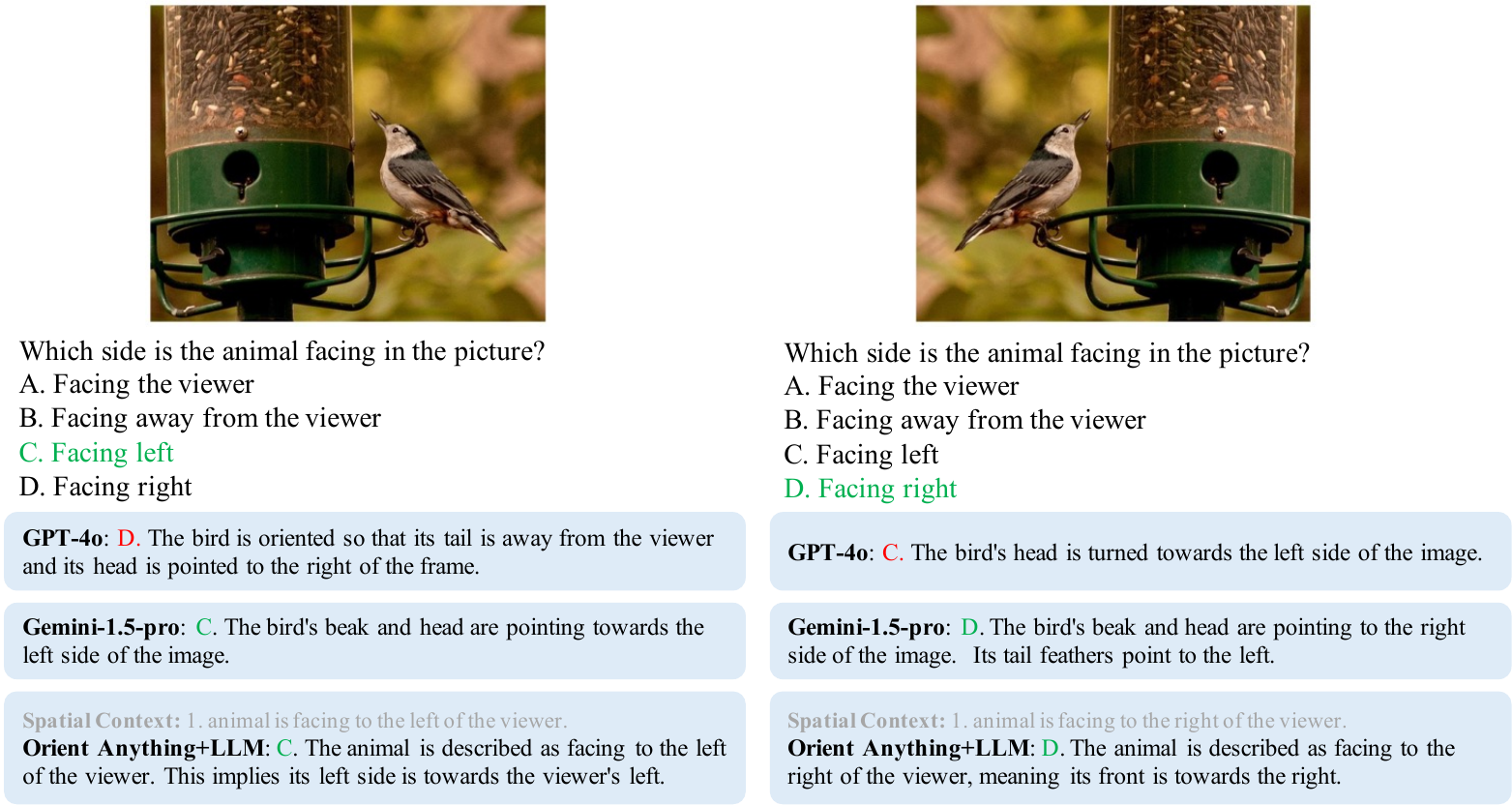}
    \includegraphics[width=0.7\linewidth]{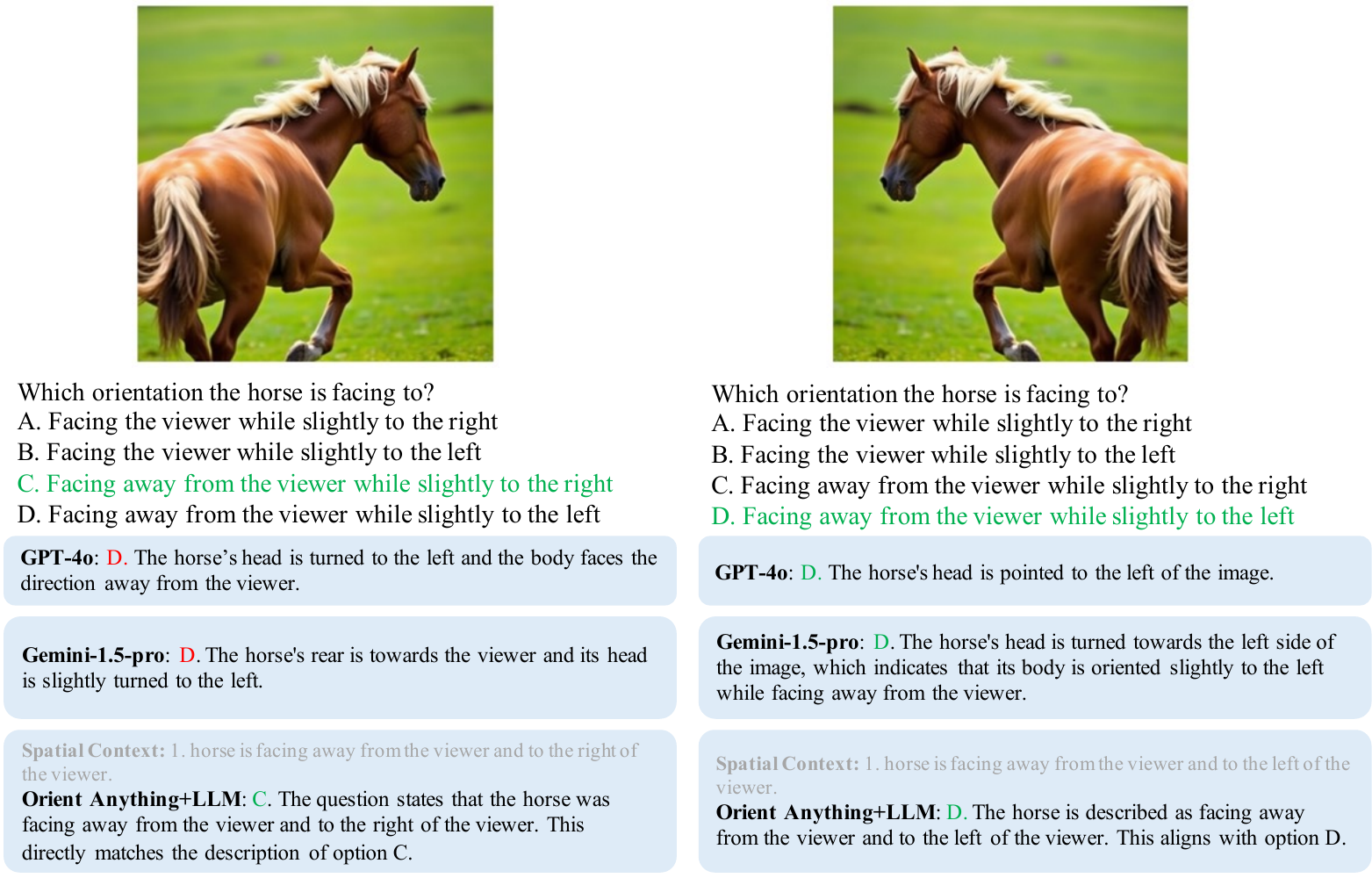}
    \includegraphics[width=0.7\linewidth]{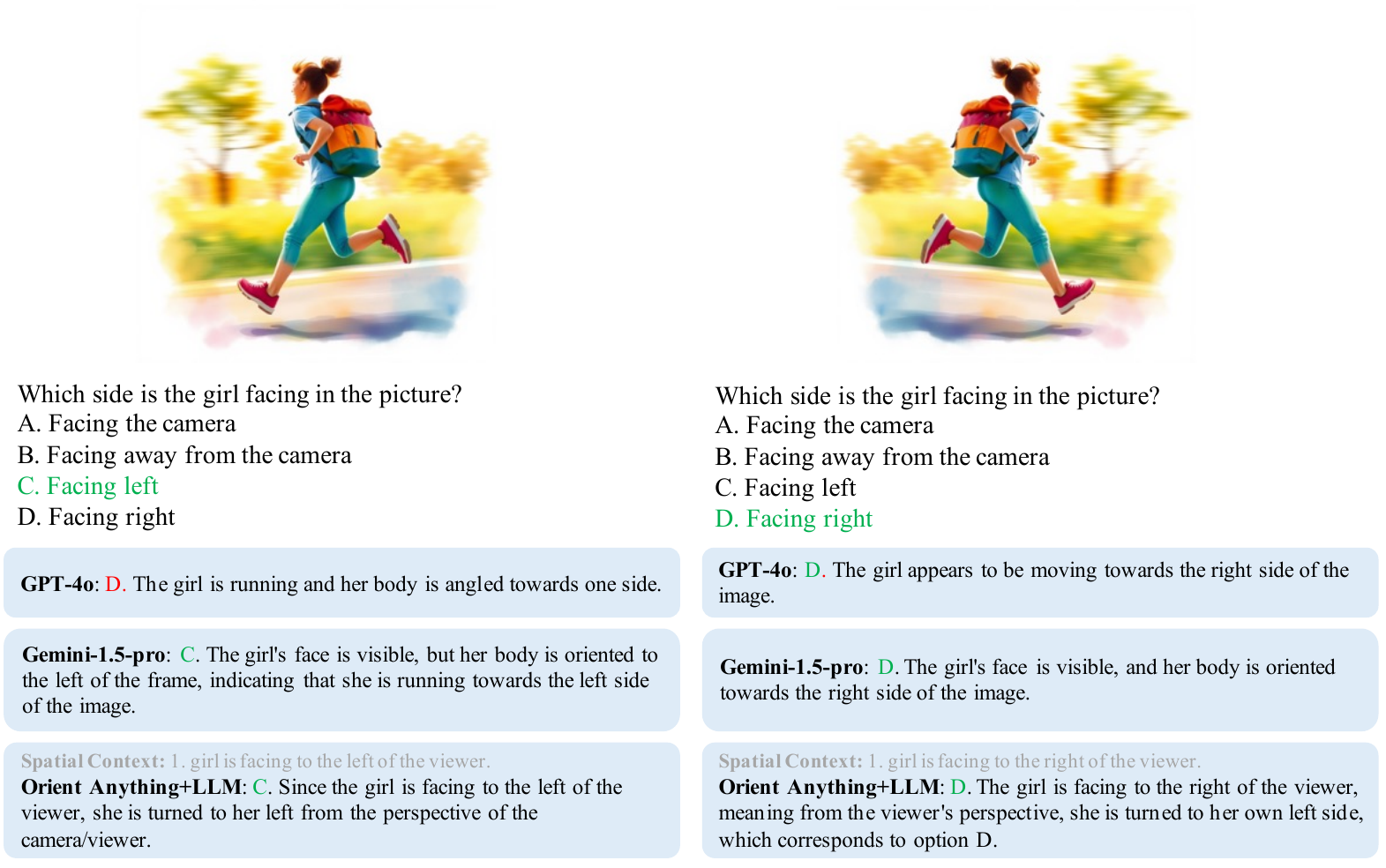}
    \caption{Visualization and qualitative comparison on the \textit{Object Direction Recognition} task of Ori-Bench.}
    \label{fig:direction}
\end{figure*}

\begin{figure*}[t]
    \centering
    \vspace{-2\baselineskip}
    \includegraphics[width=0.7\linewidth]{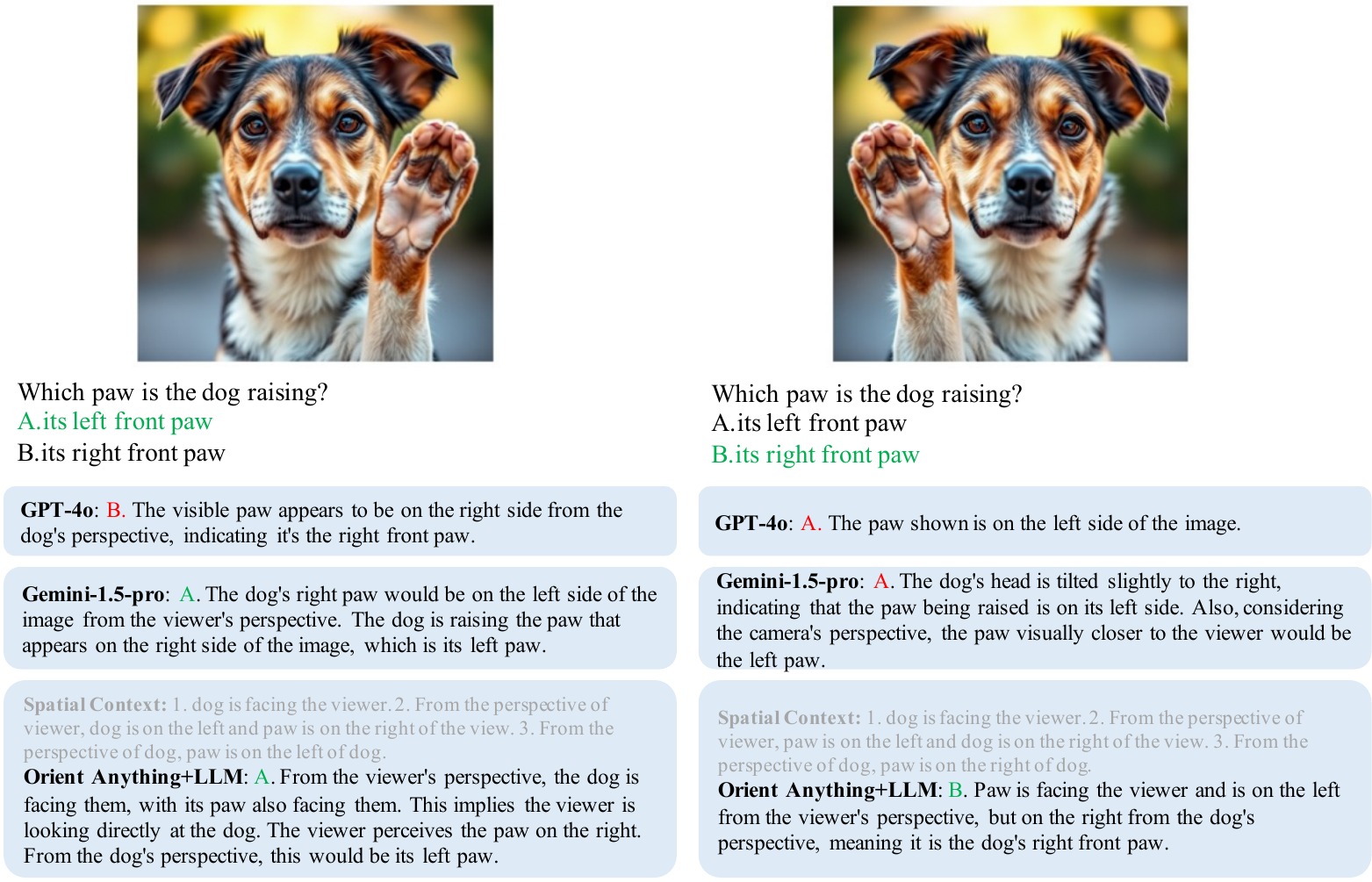}
    \includegraphics[width=0.7\linewidth]{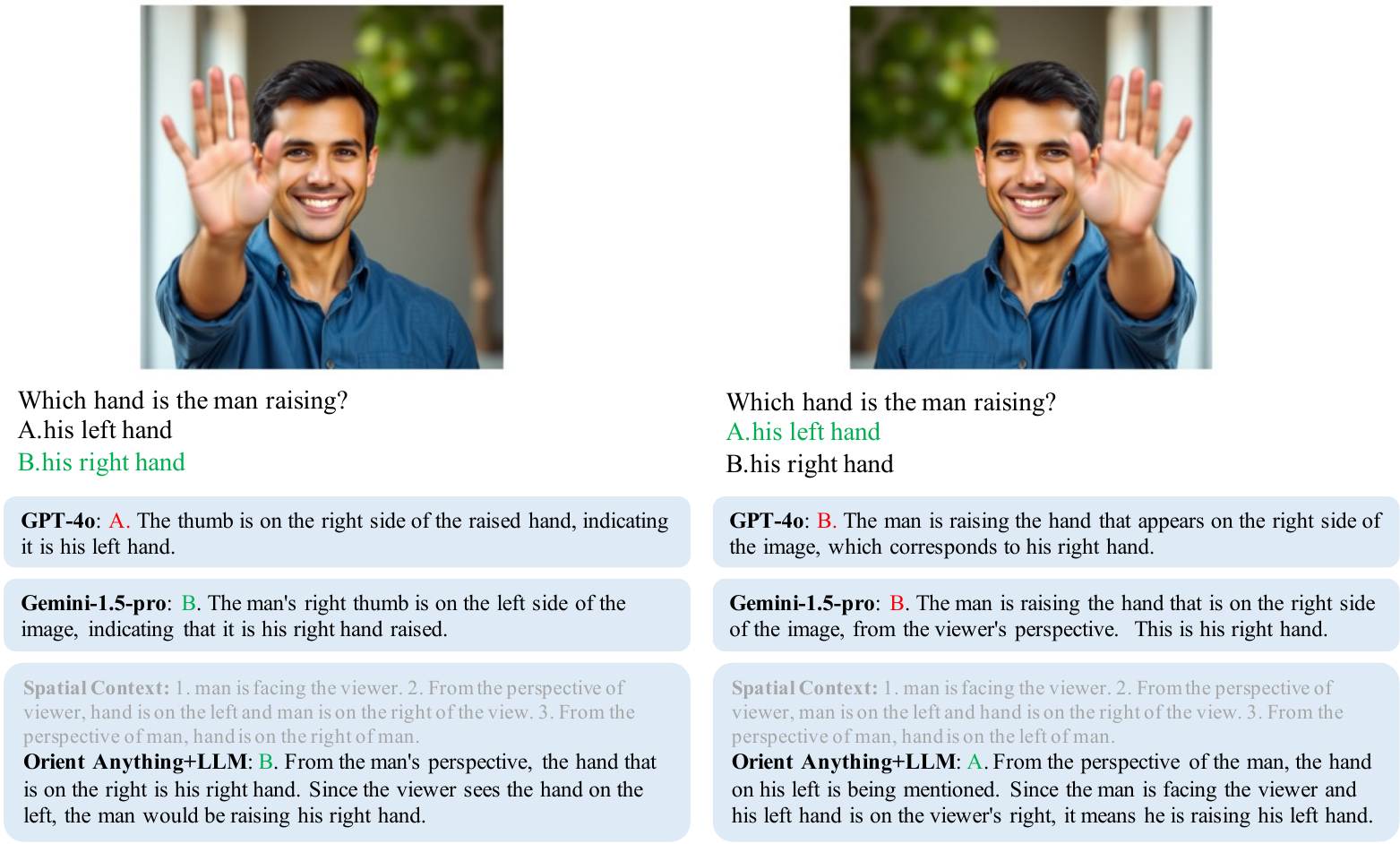}
    \includegraphics[width=0.7\linewidth]{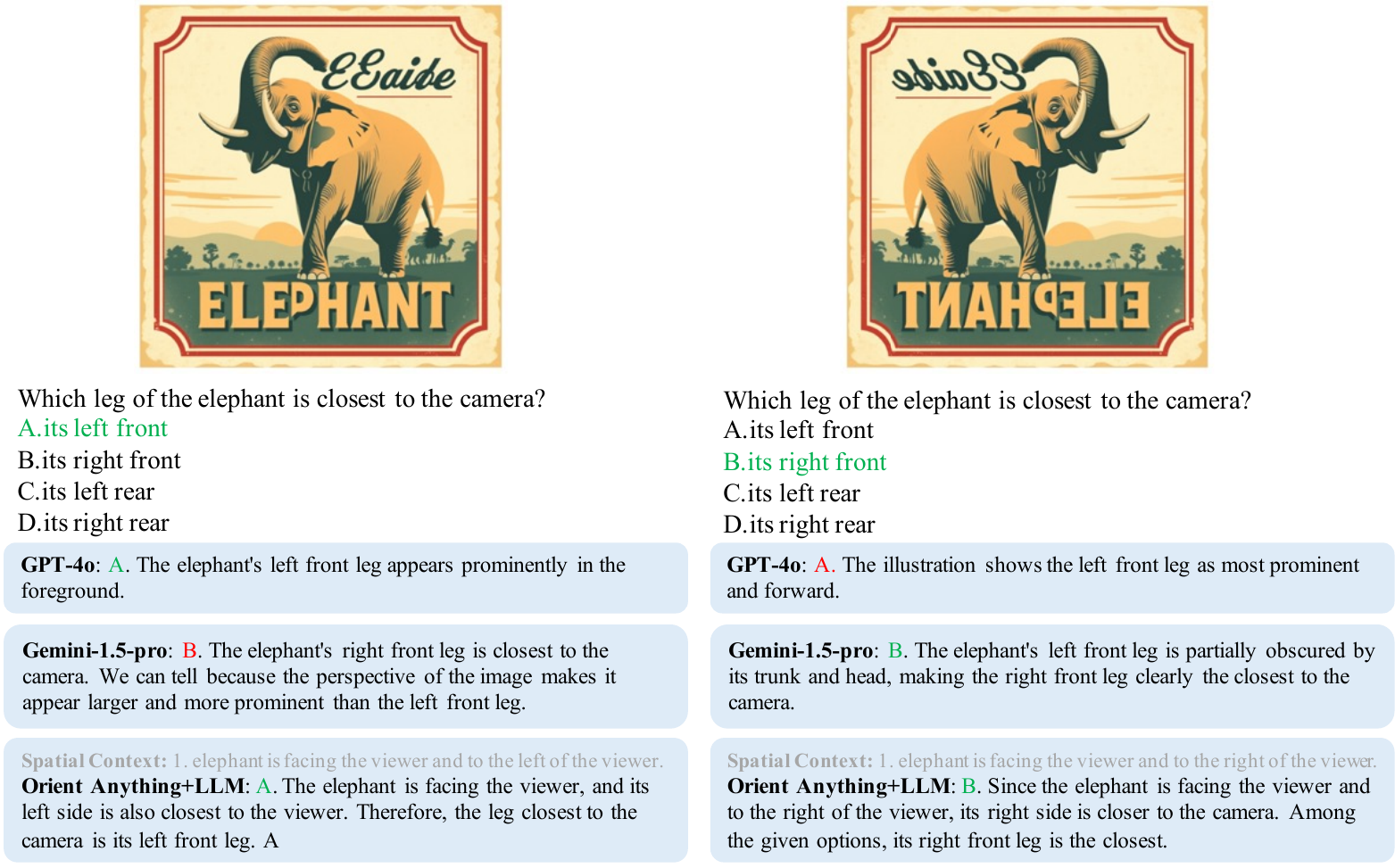}
    \caption{Visualization and qualitative comparison on the \textit{Spatial Part Reasoning} task of Ori-Bench.}
    \label{fig:part}
\end{figure*}

\begin{figure*}[t]
    \centering
    \vspace{-2\baselineskip}
    \includegraphics[width=0.7\linewidth]{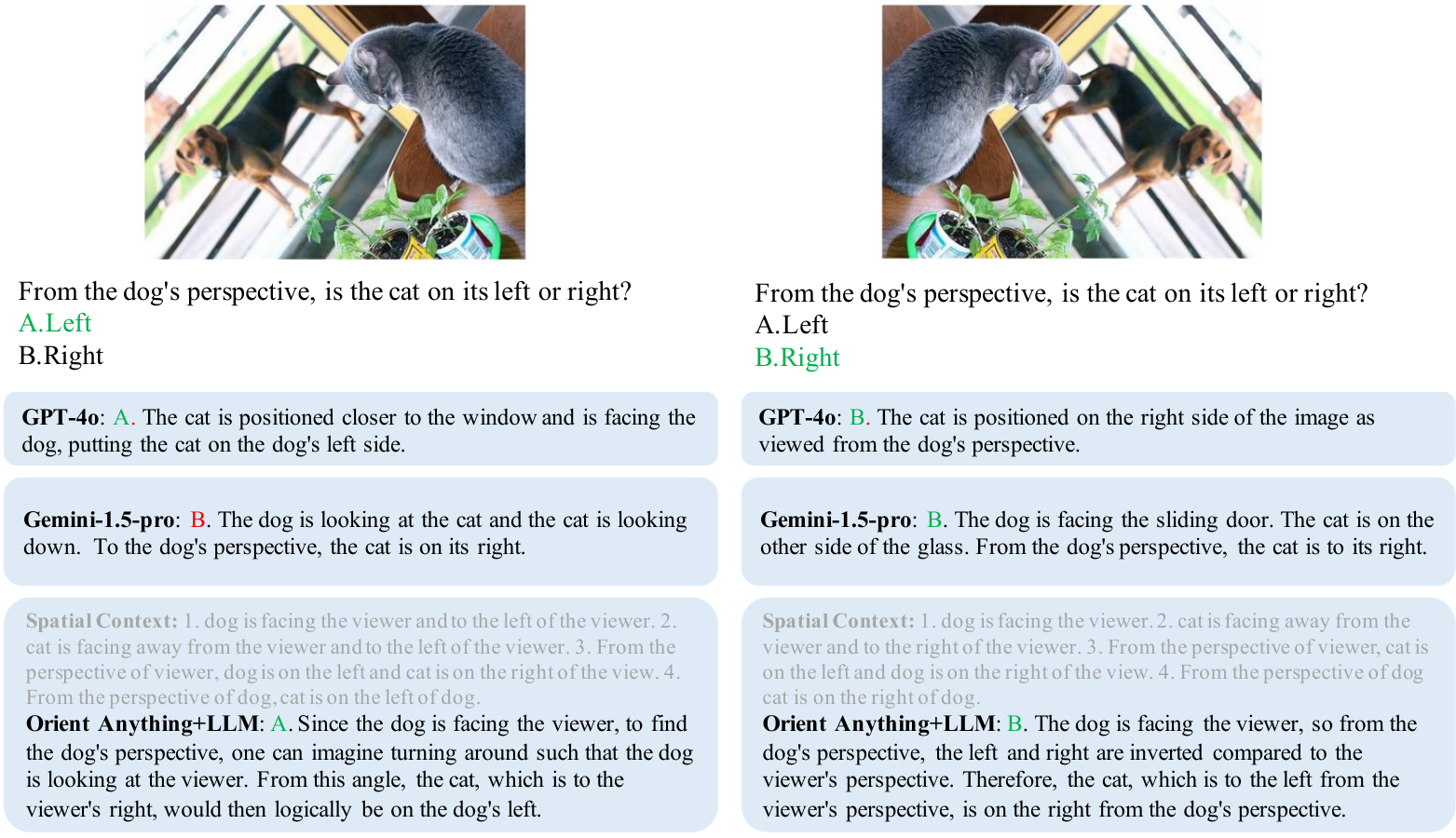}
    \includegraphics[width=0.7\linewidth]{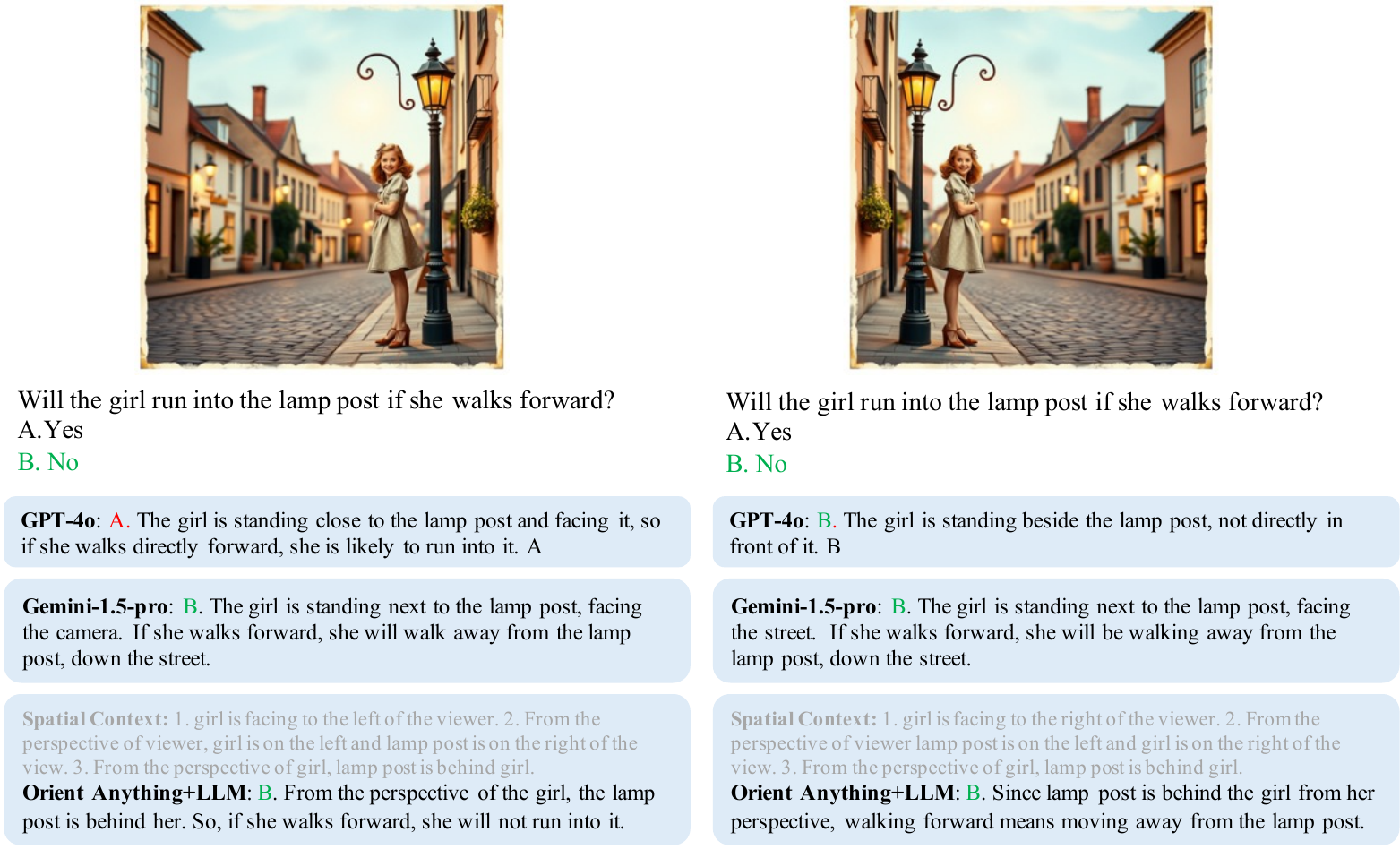}
    \includegraphics[width=0.7\linewidth]{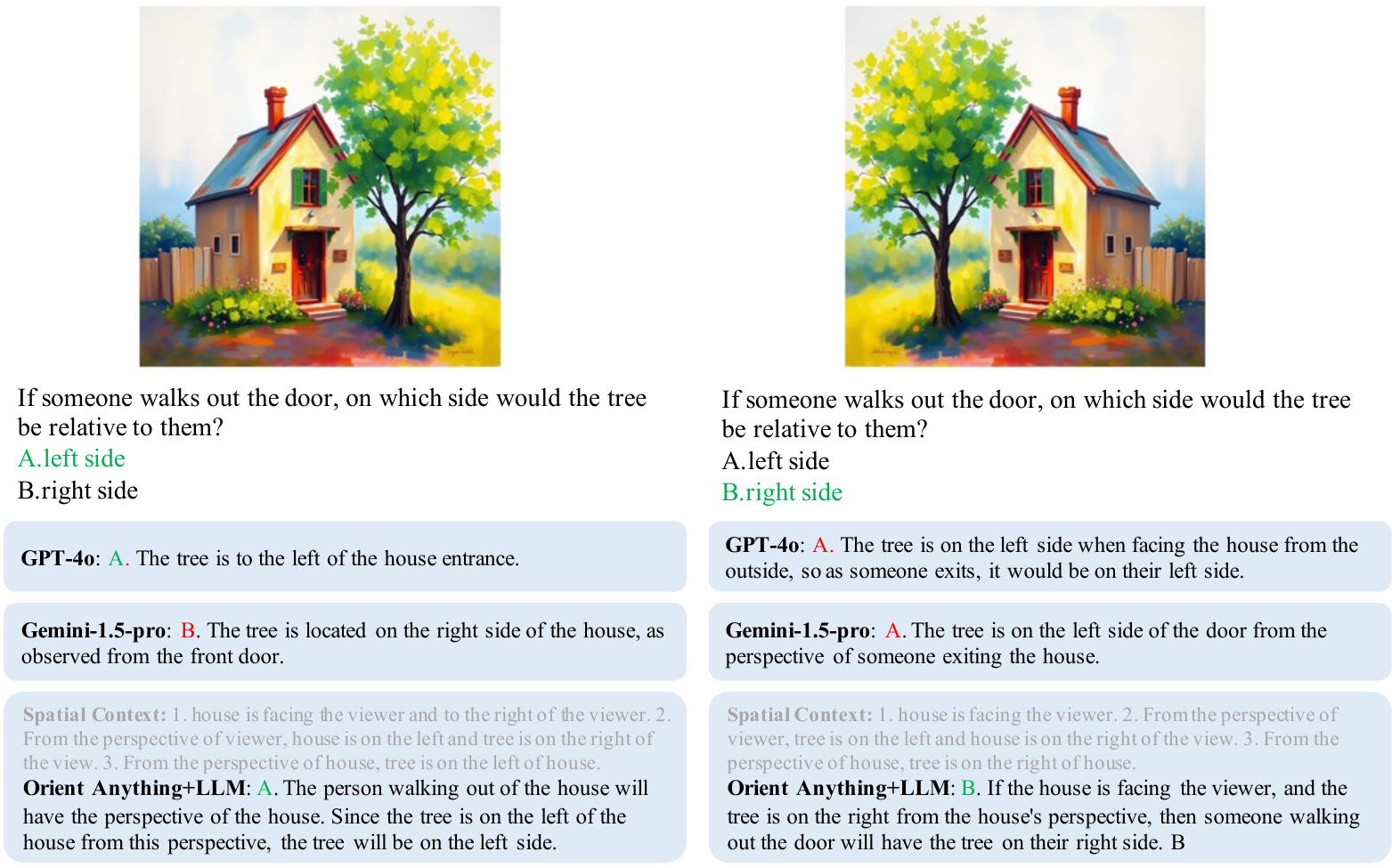}
    \caption{Visualization and qualitative comparison on the \textit{Spatial Relation Reasoning} task of Ori-Bench.}
    \label{fig:relation}
\end{figure*}

All Ori-Bench samples, along with the responses from GPT-4o, Gemini-1.5-pro, and Orient Anything+LLM, are included in the attached file. We visualize the three kinds of subtasks in Ori-Bench in Fig.~\ref{fig:direction}, \ref{fig:part} and \ref{fig:relation}, respectively.

Our observations reveal that these questions, which are intuitive for humans, often confuse the state-of-the-art VLM models like GPT-4o and Gemini-1.5-pro. This highlights the inherent limitations of existing approaches to understanding orientation. By utilizing the simple template to describe object orientations estimated by Orient Anything to LLM, we outperform alternative methods by a substantial margin.

\section{Orient Anything for Orientation Understanding}
In Section 7.1 of the main text, we briefly introduce the use of Orient Anything for solving orientation understanding problems. Here, we provide a detailed implementation.

For the open domain orientation understanding problem, we first use LLM to extract the object nouns in the question, then use Grounding-SAM~\cite{ren2024grounded} to determine the coordinates of each object, and use Orient Anything to predict the horizontal orientation of each object. 
We convert the detected spatial information into text descriptions with simple templates. For multiple objects, we use their coordinates to express their left-right relationship in the image. For each object, we only consider the azimuth angle and convert it into the horizontal 8-direction description. Finally, we provide these templated spatial descriptions, questions, and options in LLM. Practical examples are provided in Fig.~\ref{fig:direction}, \ref{fig:part}, and \ref{fig:relation}.

Although this method has obvious disadvantages (ignoring depth and 3D object relationships), it still performs much better than the Gemini-1.5-pro and GPT-4o.

The template description of the object relationship is:

\noindent  \resizebox{\linewidth}{!}{
\begin{tcolorbox}[colback=gray!10,
                  colframe=black,
                  width=\linewidth,
                  arc=1mm, auto outer arc,
                  boxrule=0.5pt,
                  box align=center,
                 ]
                 
                For \textbf{OBJ1} located in [x1, y1] with predicted azimuth angle $\hat{\varphi}$ and \textbf{OBJ2} located in [x2, y2], 

                \textit{  if x1 $<$ x2:}

                \ \ \ ``From the perspective of viewer $<$\textbf{OBJ1}$>$ is on 
                
                \ \ \ the left and $<$\textbf{OBJ2}$>$ is on the right of the view."

                \ \ \ \textit{  if 292.5° $<\hat{\varphi}<$ 360° or  0° $<\hat{\varphi}<$ 67.5°:}
                
                \ \ \ \ \ \ ``$<$\textbf{OBJ2}$>$ is on the left of $<$\textbf{OBJ1}$>$."
                
                \ \ \ \textit{  if 67.5° $<\hat{\varphi}<$ 112.5°:}
                
                \ \ \ \ \ \ ``$<$\textbf{OBJ2}$>$ is behind $<$\textbf{OBJ1}$>$."
                
                \ \ \ \textit{  if 112.5° $<\hat{\varphi}<$ 247.5°:}
                
                \ \ \ \ \ \ ``$<$\textbf{OBJ2}$>$ is on the right of $<$\textbf{OBJ1}$>$."
                
                \ \ \ \textit{  if 247.5° $<\hat{\varphi}<$ 292.5°:}
                
                \ \ \ \ \ \ ``$<$\textbf{OBJ2}$>$ is in front of $<$\textbf{OBJ1}$>$."

                 \textit{  if x1 $>$ x2:}

                \ \ \ ``From the perspective of viewer $<$\textbf{OBJ2}$>$ is on 
                
                \ \ \ the left and $<$\textbf{OBJ1}$>$ is on the right of the view."

                \ \ \ \textit{  if 292.5° $<\hat{\varphi}<$ 360° or  0° $<\hat{\varphi}<$ 67.5°:}
                
                \ \ \ \ \ \ ``$<$\textbf{OBJ2}$>$ is on the right of $<$\textbf{OBJ1}$>$."
                
                \ \ \ \textit{  if 67.5° $<\hat{\varphi}<$ 112.5°:}
                
                \ \ \ \ \ \ ``$<$\textbf{OBJ2}$>$ is in front of $<$\textbf{OBJ1}$>$."
                
                \ \ \ \textit{  if 112.5° $<\hat{\varphi}<$ 247.5°:}
                
                \ \ \ \ \ \ ``$<$\textbf{OBJ2}$>$ is on the left of $<$\textbf{OBJ1}$>$."
                
                \ \ \ \textit{  if 247.5° $<\hat{\varphi}<$ 292.5°:}
                
                \ \ \ \ \ \ ``$<$\textbf{OBJ2}$>$ is behind $<$\textbf{OBJ1}$>$."
                
\end{tcolorbox}}

The template description of object direction is:

\noindent  \resizebox{\linewidth}{!}{
\begin{tcolorbox}[colback=gray!10,
                  colframe=black,
                  width=\linewidth,
                  arc=1mm, auto outer arc,
                  boxrule=0.5pt,
                  box align=center,
                 ]
                 For \textbf{OBJ} with predicted azimuth angle $\hat{\varphi}$
                 
                 \textit{  if 292.5° $<\hat{\varphi}<$ 360° or  0° $<\hat{\varphi}<$ 22.5°:}

                \ \ \ ``The $<$\textbf{OBJ}$>$ is facing the viewer."

                 \textit{  if 22.5° $<\hat{\varphi}<$ 67.5°:}

                \ \ \ ``The $<$\textbf{OBJ}$>$ is facing the viewer and to the left of the viewer."

                \textit{  if 67.5° $<\hat{\varphi}<$ 112.5°:}

                \ \ \ ``The $<$\textbf{OBJ}$>$ is facing to the left of the viewer."
                
                \textit{  if 112.5° $<\hat{\varphi}<$ 157.5°:}

                \ \ \ ``The $<$\textbf{OBJ}$>$ is facing away from the viewer and to the left of the viewer."

                \textit{  if 157.5° $<\hat{\varphi}<$ 202.5°:}

                \ \ \ ``The $<$\textbf{OBJ}$>$ is facing away from the viewer."

                \textit{  if 202.5° $<\hat{\varphi}<$ 247.5°:}

                \ \ \ ``The $<$\textbf{OBJ}$>$ is facing away from the viewer and to the right of the viewer."

                \textit{  if 247.5° $<\hat{\varphi}<$ 292.5°:}

                \ \ \ ``The $<$\textbf{OBJ}$>$ is facing to the right of the viewer."

                \textit{  if 292.5° $<\hat{\varphi}<$ 337.5°:}

                \ \ \ ``The $<$\textbf{OBJ}$>$ is facing the viewer and to the right of the viewer."

\end{tcolorbox}}

\section{Prompts for VLMs}
\noindent \resizebox{1\linewidth}{!}{
\begin{tcolorbox}[colback=gray!10,
                  colframe=black,
                  width=\linewidth,
                  arc=1mm, auto outer arc,
                  boxrule=0.5pt,
                  box align=center,
                 ]
                 \textbf{Question Answering for Ori-Bench and Orientation Recognition}: I will ask you a single-choice question about the content of the picture. Here is the question: $<$\textbf{image}$>$ $<$\textbf{question}$>$ $<$\textbf{options}$>$.
\end{tcolorbox}}

\vspace{\baselineskip}

\noindent \resizebox{1\linewidth}{!}{
\begin{tcolorbox}[colback=gray!10,
                  colframe=black,
                  width=\linewidth,
                  arc=1mm, auto outer arc,
                  boxrule=0.5pt,
                  box align=center,
                 ]
                 \textbf{Orientation Annotating for Orthogonal Rendering views}: I'm going to show four images of the same object from four viewpoints in turn and label them `A.' `B.' `C.' `D.' Four options. The option `E.' is "No front face or More than One front Face". Decide whether it have a front and if yes, which one is the front of the object after the presentation. Note that: If the object is a gun, bow and arrow, etc., please use the muzzle of the gun as the front. Stick tools and weapons such as swords, axes, knives, and wrenches are considered to have no front. If you cannot decide or there is more than one front, you should choose `E.'. A.$<$\textbf{image viewA}$>$ B.$<$\textbf{image viewB}$>$ C.$<$\textbf{image viewC}$>$ D.$<$\textbf{image viewD}$>$ E.No front face.
\end{tcolorbox}}

\vspace{\baselineskip}

\noindent \resizebox{1\linewidth}{!}{
\begin{tcolorbox}[colback=gray!10,
                  colframe=black,
                  width=\linewidth,
                  arc=1mm, auto outer arc,
                  boxrule=0.5pt,
                  box align=center,
                 ]
                 \textbf{Accurate Orientation Angles Estimation}: I will ask you a question about the content of the picture. Here is the question: $<$\textbf{image}$>$ Align the front of the object towards the viewer. Rotate the object x degrees to its right (i.e., clockwise from a top view), using a 360° per full circle unit system. Adjust the height of the viewer to form a pitch Angle y with the object (same unit of degrees; y is positive if the viewer is looking down at the object, and y is negative if the viewer is looking up at the object). Finally, the viewer is rotated clockwise by an Angle z (same unit of degrees) with the line connecting the viewer and the object as the axis, and a negative z indicates a counterclockwise rotation. Now, please directly predict the values of x, y, and z in float format.
\end{tcolorbox}}

\end{document}